\documentclass[10pt,twocolumn,letterpaper]{article}

\usepackage{iccv}
\usepackage{times}
\usepackage{epsfig}
\usepackage{graphicx}
\usepackage{amsmath}
\usepackage{amssymb}
\usepackage{pifont}
\usepackage{multirow}
\usepackage{tabu}
\usepackage{multicol}
\usepackage{booktabs}
\usepackage{wrapfig}
\usepackage{graphicx}
\usepackage{subcaption}
\usepackage{adjustbox}
\usepackage{makecell}
\usepackage[normalem]{ulem}

\usepackage{tikz}
\usepackage{pgfplots}
\pgfplotsset{compat=default}
\usetikzlibrary{pgfplots.groupplots}

\usepackage[pagebackref=true,breaklinks=true,letterpaper=true,colorlinks,bookmarks=false]{hyperref}

\usepackage[capitalize]{cleveref}
\crefname{section}{Sec.}{Secs.}
\Crefname{section}{Section}{Sections}
\Crefname{table}{Table}{Tables}
\crefname{table}{Tab.}{Tabs.}

\makeatletter
\DeclareRobustCommand\onedot{\futurelet\@let@token\@onedot}
\def\@onedot{\ifx\@let@token.\else.\null\fi\xspace}

\def\eg{\emph{e.g}\onedot}%\def\Eg{\emph{E.g}\onedot}

\def\ie{\emph{i.e}\onedot}
\def\cf{\emph{cf}\onedot}% \def\Cf{\emph{Cf}\onedot}

\def\vs{\emph{vs}\onedot}

\makeatother

\iccvfinalcopy % *** Uncomment this line for the final submission

\def\iccvPaperID{8972} % *** Enter the ICCV Paper ID here

\ificcvfinal\fi

\begin{document}

%%%%%%%%% TITLE
\title{Multiscale Memory Comparator Transformer for Few-Shot Video Segmentation}

\author{Mennatullah Siam, Rezaul Karim, He Zhao, Richard Wildes\\
  Department of Electrical Engineering and Computer Science\\
  York University\\
  Toronto, CA \\
  \texttt{\{msiam, zhufl, wildes\}@eecs.yorku.ca, karimr31@yorku.ca} \\
}

\maketitle
\ificcvfinal\thispagestyle{empty}\fi

\begin{abstract}
%Learning to compare support and query features in few-shot video segmentation was shown to be a powerful approach. 
Few-shot video segmentation is the task of delineating a specific novel class in a query video using few labelled support images.
Typical approaches compare support and query features while limiting comparisons to a single feature layer and thereby ignore potentially valuable information.
We present a meta-learned Multiscale Memory Comparator (MMC) for few-shot video segmentation that combines information across scales within a transformer decoder. Typical multiscale transformer decoders for segmentation tasks learn a compressed representation, their queries, through information exchange across scales. Unlike previous work, we instead preserve the detailed feature maps during across scale information exchange via a multiscale memory transformer decoding to reduce confusion between the background and novel class. Integral to the approach, we investigate multiple forms of information exchange across scales in different tasks and provide insights with empirical evidence on which to use in each task. The overall comparisons among query and support features benefit from both rich semantics and precise localization. We demonstrate our approach primarily on few-shot video object segmentation and an adapted version on the fully supervised counterpart. 
%To further show our generality, we also extend the approach to actor/action segmentation. 
In all cases, our approach outperforms the baseline and yields state-of-the-art performance. Our code is publicly available at \url{https://github.com/MSiam/MMC-MultiscaleMemory}.

%: Comparators that operate with early network layer features support precise localization, but lack sufficient semantic abstraction; at the other extreme, operating with deeper layer features provides richer descriptors, but sacrifices localization. 
%We address this scale selection challenge with 
\end{abstract}

\section{Introduction}
\begin{figure}[t]
    \centering
       \includegraphics[width=0.45\textwidth]{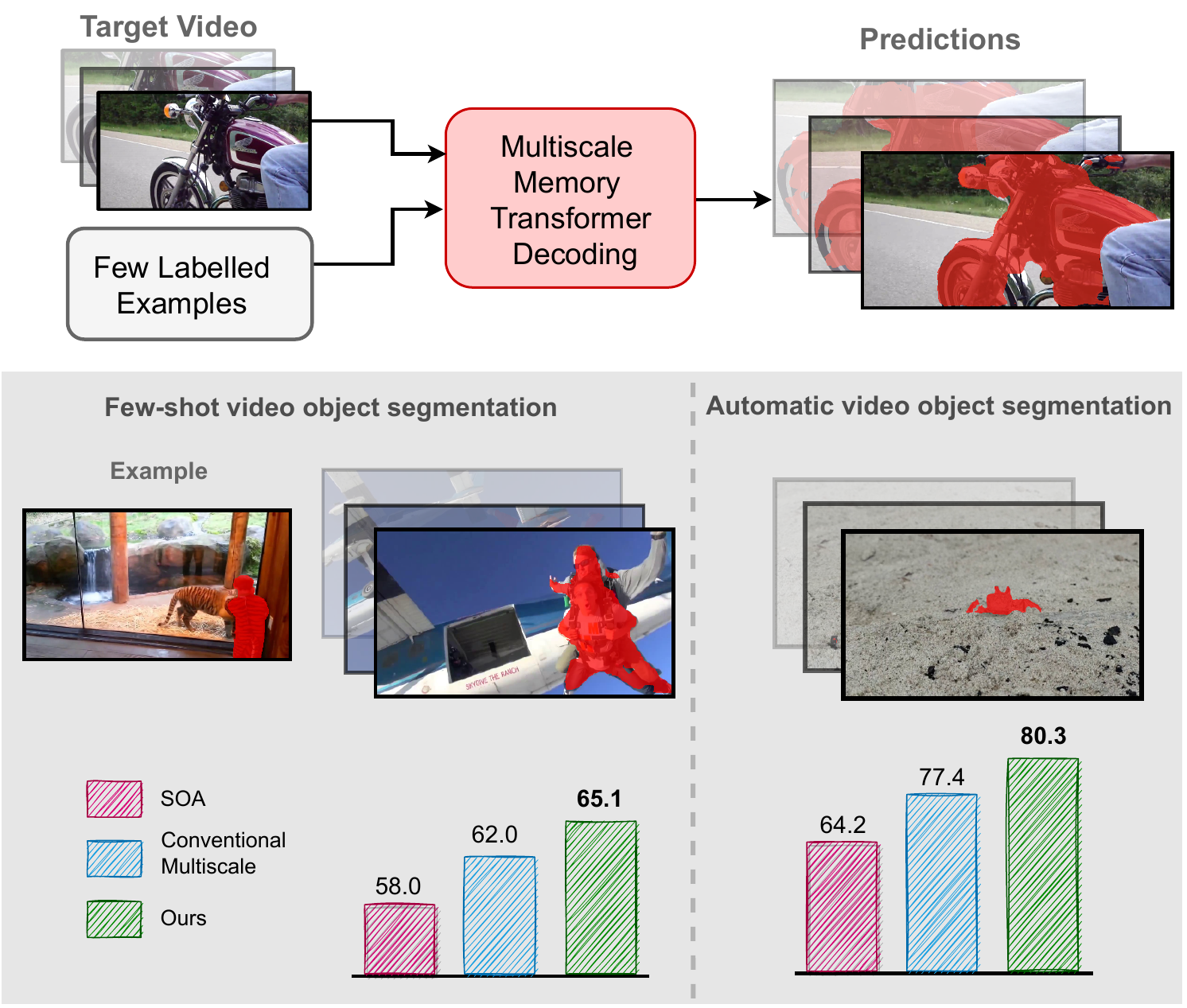}
    \captionsetup{labelfont=bf}
    \caption{We present Multiscale Memory Comparator Transformer, MMC Transformer, for FS-VOS, which is the first multiscale video transformer to compare few labelled examples and a target video. \textbf{Top:} At the heart of our method is a multiscale memory transformer decoder for support-query comparisons that preserves the spatiotemporal features during multiscale processing instead of relying on a compact representation. \textbf{Bottom:} We evaluate primarily on FS-VOS and its fully supervised counterpart. MMC Transformer (ours) outperforms the conventional multiscale baselines and state of the art (SOA) on these tasks.}
    \label{fig:overview}
    \vspace{-1.5em}
\end{figure}

% What & Why
Guided by a few labelled examples (\ie the support set), few-shot learning is focused on improving the generalization ability of models to novel classes unseen during the initial training to classify the query images. Most approaches emulate the inference stage during training by sampling tasks of support and query sets~\cite{Triantafillou2020Meta-Dataset}. More generally, this approach is referred to as \textit{meta-learning}~\cite{metalearn_tutorial}.
%\textcolor[red}{[REVIEW CITE]. More generally, his approach of emulating inference during training is referred to as \textit{meta-learning} [CANONICAl SITE]}. 
On the other hand, video segmentation~\cite{wang2021survey} is concerned with exploiting  temporal structure in the data when delineating different objects in a video. In this paper, we focus on the intersection of meta-learning and few-shot video object segmentation (FS-VOS). We specifically focus on learning support-query comparisons (i.e. comparators), with temporal consistency. 
%these two using metric learning for few-shot video object segmentation (FS-VOS)  \textcolor{blue}{within a meta-learning framework}. Thus, learning to compare features between the support and query video sets (\ie learning comparators), while considering the temporal structure. \textcolor{red}{RW: I've commented the last two sentences out, because everything before was seeting-up FS, VOS and meta-learning, but then at the very end you introduce apparently new concepts of metric learning and compaerators. If you want to do that, then you need to intergrate them with everything that came before; otherwise, its very confusing.}

Previous FS-VOS work has documented that pixel-to-pixel comparisons between the query and support sets better capture fine details compared to working with global average pooled representations~\cite{chen2021delving,siam2020weakly}. Still, these efforts  did not explore multiscale comparisons. A key question arises: Which features should be compared?  Limiting comparisons between the support and query sets at the finest scales (\ie shallow network layers) capture only primitive semantics (\eg local orientation) and thus are error prone, while the coarser scales (\ie deeper network layers) capture abstract semantics but sacrifice detailed information that support precise localization. In response we propose a multiscale comparator for few-shot video dense prediction tasks, MMC Transformer. Previously, few-shot image segmentation using correlation tensors between features from multiple scales has been investigated~\cite{min2021hypercorrelation}. However, its use of 4D convolution and squeeze operations on potentially erroneous correlations can lead to confusion between the novel class and the background. It also did not provide a means to ensure the temporal consistency of the extracted features, since it was designed for images not videos. For fully supervised panoptic segmenation, a multiscale transformer decoder has been presented as a means to improve the separation across the different categories/instances\cite{cheng2021masked}; however, its multiscale decoding lead to the loss of the spatiotemporal dimension during the multiscale processing, which can compromise the segmentation.

%How
To address the scale selection challenge with better separation of the novel class and the background, we present a novel Multiscale Memory Comparator (MMC) Transformer that takes as input a set of squeezed correlation tensors that encompass comparisons between support-query features at multiple scales. To overcome the challenge in previous multiscale transformer decoders that pooled the information into a compact vector~\cite{cheng2021masked,cheng2021mask2former}, we propose multiscale memory decoding. Our design allows the multiscale processing within the transformer decoder to preserve the spatiotemporal dimension and to provide temporally consistent attention maps from the meta-learned memory entries. For the challenge of handling erroneous correlations, we propose a bidirectional information exchange across scales, (\ie coarse-to-fine and fine-to-coarse), inspired by multigrid methods~\cite{briggs2000multigrid}. Critically, this exchange reduces the impact of erroneous correlations at the finer scales by incorporating feedback from coarser scales and vice versa. These innovations are embedded in a meta-learning framework to exploit task structure in learning memory entries that can segment the novel class guided by the support-query comparisons.
%\textcolor{red}{Possibly move contributions after related work.}\textcolor{blue}{ Do you prefer I switch the related work back to the intro? Since usually contributions are in the introduction to help the readers grasp early on what we are working on. Since our previous round of reviews had issues with the novelty I am suggesting to keep it in the intro.}

{\bf Contributions.} In summary, our main contributions are threefold: (1) We present the first meta-learned multiscale comparator between the support and query in few-shot video dense prediction tasks, with state-of-the-art performance.
%SOA performance is part of our contributions so is it ok if we keep it but I reduced it for space?
(2) We present multiscale memory transformer decoding that performs information exchange across scales on the dense feature maps to capture detailed information, rather than an overly compact representation that sacrifices such distinguishing detail. This decoding scheme is used in our comparator transformer.
%in a meta-learning framework that operates on the correlation feature pyramid to better separate the novel class and the background. Our multiscale memory decoding yields temporally consistent attention maps in relation to the memory entries.
%and improves on both few-shot and fully supervised video object segmentation baselines.}
(3) We investigate different forms of information exchange across scales. This study motivates a final comparator design that operates with bidirectional multiscale information exchange. 
%(4) We propose transformer based boundary refinement with separate segmentation and boundary prediction heads that input the multiscale memory decoded features and the original query features. 
Our approach outperforms a multiscale baseline and the state of the art on FS-VOS and automatic video object segmentation (AVOS); see Fig.~\ref{fig:overview}.  
%\textcolor{red}{Two things of concern: (1) Some will object that its inappropriate to confine a new task to the supplement, because it is more than supplemental. (2) Even if a reviewer does not outright object, there is no guarantee they will look at; so, impact may be minimal.} 
%Additionally, we compare our multiscale memory learning with respect to the baseline on fully supervised video object/action segmentation. Our code will be publicly released upon acceptance.

\section{Related work}
%\textcolor{blue}{reduce in FSL and VOS 1 sentence each.}
\textbf{Few-shot learning.} Metric learning (\ie learning to compare) is widely adopted in few-shot 
classification (\eg ~\cite{qi2018low,snell2017prototypical}), 
segmentation (\eg ~\cite{wang2019panet,siam2020weakly,min2021hypercorrelation}), video object segmentation~\cite{chen2021delving} and action localization in video~\cite{yang2021few}. Multiscale processing often is not exploited in few-shot video tasks~\cite{chen2021delving,yang2021few}, we are the first to exploit multiscale support-query comparisons in dense few-shot video tasks. While work in few-shot segmentation has considered multiscale processing in comparators~\cite{min2021hypercorrelation}, the model can be confused by erroneous correlations. In our work, we investigate meta-learning a multiscale memory transformer decoding that better separates the background and novel class by allowing bidirectional exchange across scales. Moreover, it provides temporally consistent attention maps.
%, especially at the finest scales

\textbf{Video object segmentation.} Deep video object segmentation (VOS) approaches~\cite{wang2021survey} can be categorized into automatic and semi-automatic methods. Automatic VOS (AVOS) segments the primary objects in videos without a predefined initialization, while semi-automatic VOS requires first frame mask initialization to consequently track objects. We illustrate the generalizability of our approach on AVOS since it is more challenging, as no initialization is provided. The current state of the art on AVOS uses both optical flow and RGB images as input with traditional coarse-to-fine convolutional decoding schemes~\cite{zhou2020motion, ren2021reciprocal, jain2017fusionseg}. In contrast, our model uses multiscale memory transformer decoding that takes an input clip without optical flow.

\begin{figure*}[t]
    \centering
       \includegraphics[width=0.9\textwidth]{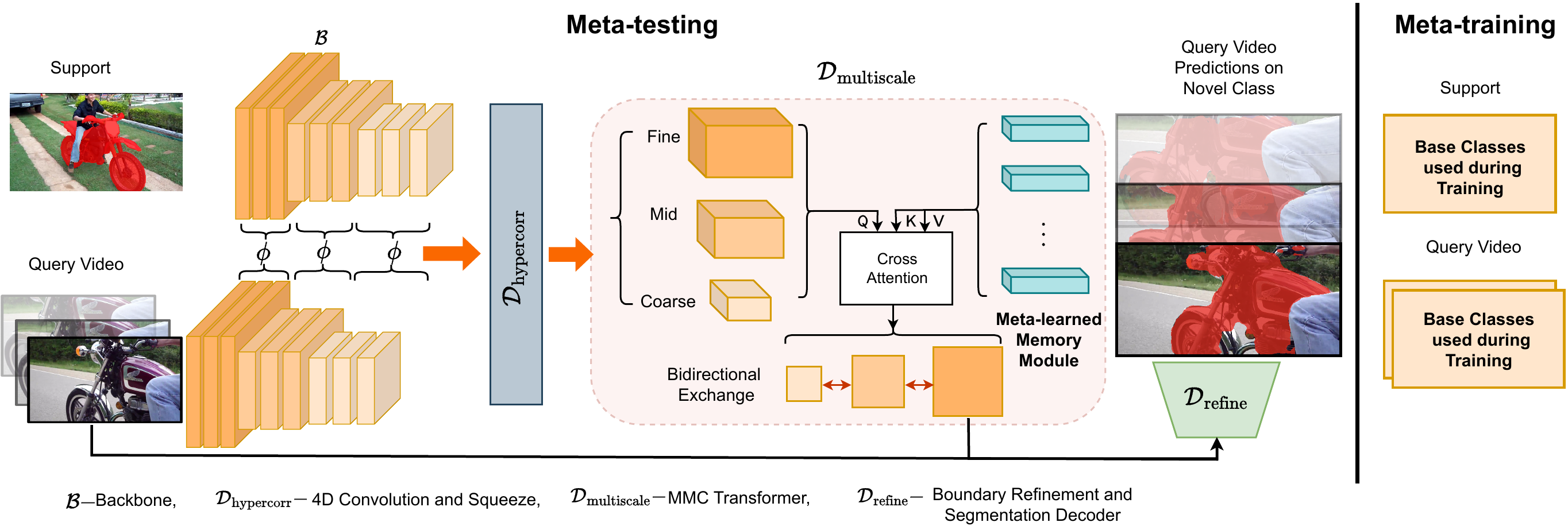}
    \vspace{-1em}
    \captionsetup{labelfont=bf}
    \caption{Overview of our FS-VOS architecture. Features from a convolutional backbone, $\mathcal{B}$, are extracted. 4D correlation tensors are computed among support-query features, with $\phi$ symbolizing correlation. Subsequently, the tensors go through 4D convolution and squeeze operations, $\mathcal{D_{\text{hypercorr}}}$, that yield 2D feature maps for memory efficiency. A multiscale memory comparator transformer, $\mathcal{D}_{\text{multiscale}}$, operates on the input pyramid.
    %$\{\bar{\mathsf{X}}_l\}_{l=1}^L$
    Cross attention among the feature pyramid, $\mathsf{Q}$, and the learnable memory, $\mathsf{K}, \mathsf{V}$, is performed, Eq.~\ref{eq:st}. Information is transferred bidirectionally between scales (denoted by red arrows); the output at the finest scale and the target query video are input to the boundary refinement and segmentation decoder, $\mathcal{D}_{\text{refine}}$. We follow a meta-learning framework: During training we sample support-query pairs on the base classes (\ie meta-training), then evaluate on the novel class support-query sets during inference (\ie meta-testing); see Sec.~\ref{sec:metalearning_setup}.}
    %\textcolor{red}{Move this figure earlier, \ie back near where its first referred to before Sec. 3.1.}}
    \label{fig:detailed}
    \vspace{-1em}
\end{figure*}

\textbf{Multiscale transformers.} Incorporating multiscale processing in transformers is an emerging topic. Previous work has mainly explored multiscale information on the encoder side~\cite{fan2021multiscale,li2021improved,wang2022crossformer,wu2022memvit}, which is not sufficient when computing dense predictions. Recently, a multiscale transformer decoder was proposed~\cite{cheng2021masked,cheng2021mask2former}, but its design exchanged information across scales on compressed learnable queries. Our approach preserves the spatiotemporal dimension during across scale information exchange to yield detailed segmentations. Moreover, unlike previous transformers, ours operates on correlation tensors rather than feature maps to induce a stronger bias toward learning a comparator. 

Our bidirectional exchange across scales can be seen as reminiscent of multigrid methods for numerical solutions~\cite{briggs2000multigrid}, their prior application to computer vision problems (\eg\cite{terzopoulos1986,galum2015}) and specifically segmentation, \eg~\cite{sharon2006hierarchy}. Recently, it was explored in convolutional architectures~\cite{ke2017multigrid} and to accelerate its training~\cite{wu2020multigrid}, yet we are the first to investigate it in multiscale transformers. Concurrent work using the Swin-Transformer~\cite{hong2022cost} encoded correlation tensors in few-shot image segmentation. However, our work is focused on videos and is orthogonal to theirs, where we focus on multiscale transformer decoders, not encoders. Additionally, our approach is applicable to both FS-VOS and AVOS. Another concurrent work~\cite{wu2022memvit} used a memory module in the encoder to extract information across clips. Our multiscale memory is in the decoder to maintain dense feature maps, while meta-learning support-query relations.

\section{Multiscale memory transformer comparator}

%\subsection{Multiscale multigrid attention (MMA)}
%\label{sec:multigrid}

Our Multiscale Memory Comparator (MMC) Transformer centres around a novel multiscale memory decoder that operates on a spatiotemporal dimension preserving representation (\cref{sec:msmem}). Within this decoder attentional processing operates with bidirection cross-scale information exchange to combat erroneous correlations that might appear at any individual scale (\cref{sec:multigrid}). Subsequently, the output features accompanied by the original query video go through boundary refinement and a segmentation decoder to provide the final video predictions (\cref{sec:bdry_refine}).

Input features to the MMC Transformer are task dependant. In this section, we concentrate on FS-VOS; Sec.~\ref{sec:method_overview} presents the minor modifications necessary for application to AVOS. For FS-VOS the input comes as support-query correlation tensors between features from a convolutional backbone. Figure~\ref{fig:detailed} overviews the entire architecture.

\subsection{Preliminaries}
 Since our paper explores both few-shot learning and transformers, to reduce ambiguities between the term \textit{query} used in both, we use the term \textit{target query} when referring to its use in few-shot throughout the rest of the paper. \\

We subscript with $l$ to denote features from scale level $l \in \{1, 2, \dots, L\}$, which are extracted from late (coarse), $l=1$, intermediate or early (fine), $l=L$, stages. 
%\textcolor{red}{RW: Seems indexing is backward, \ie shouldn't $l=1$ be fine and $l=L$ coarse?}
The input features, $\mathsf{X}_l$, are extracted from scale level $l$. 
%\in \mathbb{R}^{TH_lW_l \times C_l}$, 
For the sake of memory efficiency, we perform dimensionality reduction with a $1\times 1$ convolutional layer followed by flattening to yield the final features at level $l$ as $\bar{\mathsf{X}}_l \in \mathbb{R}^{TH_lW_l \times D}$, where $H_l$ and $W_l$ are spatial dimensions, $T$ is the clip length and $D$ is the channel dimensions. Following standard convention, let multihead attention~\cite{vaswani2017attention}, $\mathcal{A}$, be defined as
\vspace{-0.2em}
 \begin{equation}
\hspace*{-0.2cm} \mathcal{A}(\mathsf{Q},\mathsf{K},\mathsf{V}) = \left[\oplus_{h=1}^{N_h}\bigl(\text{Softmax}\left(\frac{\mathsf{Q_h}\mathsf{K_h}^\top}{\sqrt{D}}\right)\mathsf{V_h}\bigr)\right]\mathsf{W}^o,
    \label{eq:multihead}
\end{equation}
where the inputs $\mathsf{Q}, \mathsf{K}$ and $ \mathsf{V}$ represent the query, key and value, resp., with $D$ feature dimensions and $\oplus$ the concatenation operator. The subscript $\mathsf{.}_h$ indicates the corresponding projected tensor per head and $\mathsf{W}^o$ is the learnable weights to merge the output from $N_h$ heads.
%\textcolor{blue}{The output from different heads are combined as per standard procedure~\cite{vaswani2017attention}.}

\subsection{Multiscale memory decoding}
\label{sec:msmem}

\begin{figure}[t]
\vspace{-1em}
    \centering
       \includegraphics[width=0.45\textwidth]{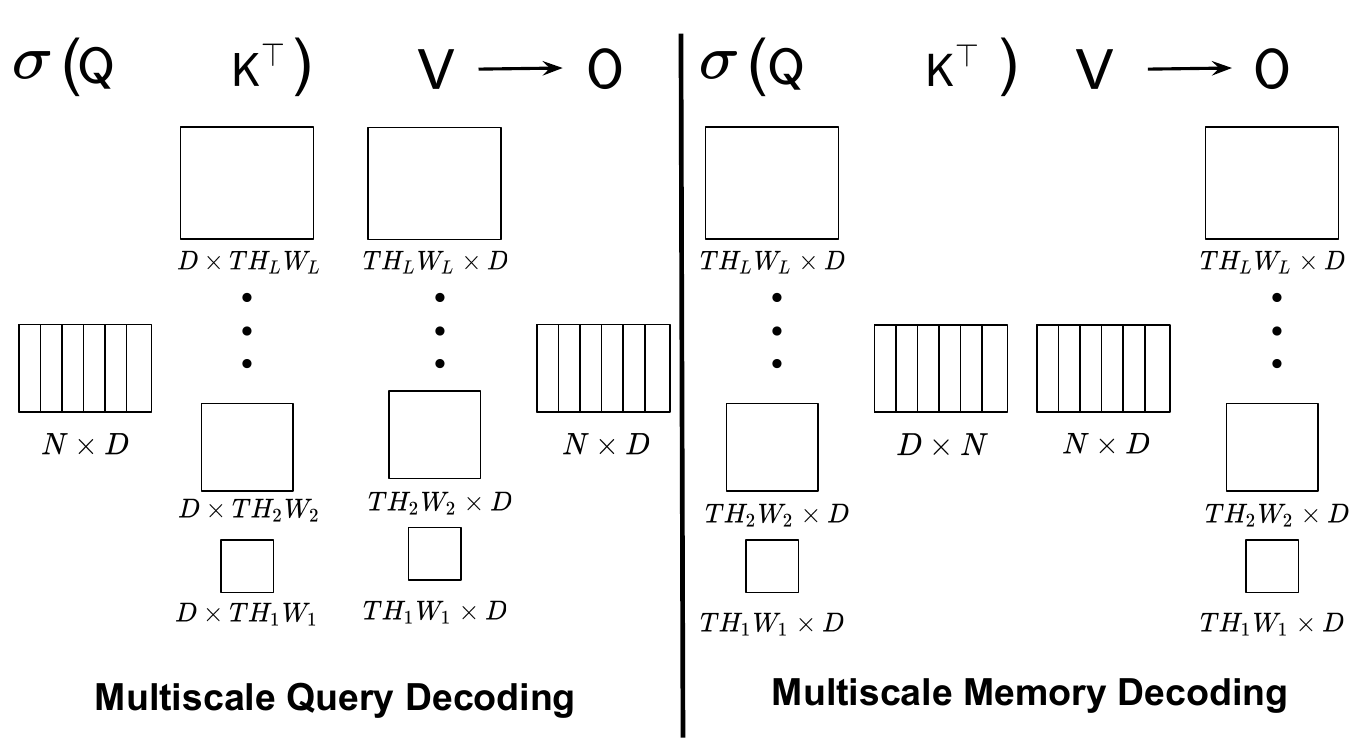} 
    \vspace{-1em}
    \captionsetup{labelfont=bf}
    \caption{Two variants of multiscale processing in transformer decoder: multiscale query decoding \vs multiscale memory decoding. $\mathsf{Q}, \mathsf{K}, \mathsf{V}$ are the inputs to the multihead attention at every scale, $\sigma$ is the softmax operator, $\mathsf{O}$ is the corresponding output, $TH_lW_l$ is the spatiotemporal dimension for scale $l \in \{1, 2, \cdots, L$\}, $D$ is the feature dimension and $N$ is the number of learnable entries. }
    \label{fig:multiscalemem}
    \vspace{-1em}
\end{figure}

A key idea in our approach is to enable  multiscale processing and information transfer across scales to capture detailed information in the feature maps during decoding. Hence, we seek a formulation that preserves the spatiotemporal dimensions, rather than one that employs a compressed representation; see  Fig.~\ref{fig:multiscalemem}.  Correspondingly, we meta-learn a memory module that has $D$ dimensional vectors, $\mathsf{M}^{\text{memory}} \in \mathbb{R}^{N\times D}$, with $N$ memory entries that are shared across all decoding layers and scales. This multiscale memory decoding allows the per-scale decoded output to preserve the spatiotemporal dimension, unlike~\cite{cheng2021masked}. We perform cross attention per resolution and feature abstraction level, $l$, where we instantiate the multihead attention, \eqref{eq:multihead}, as
\begin{equation}
    \mathsf{O}_l = \mathcal{A}(\bar{\mathsf{X}}_l + \mathsf{P}^{sc}_l + \mathsf{P}^{st}_l, \mathsf{M}^{\text{memory}} + \mathsf{P}, \mathsf{M}^{\text{memory}}),
    \label{eq:st}
\end{equation}
with $\mathsf{P} \in \mathbb{R}^{N \times D}$ learnable memory positional embeddings and $\mathsf{P}^{st}_l \in \mathbb{R}^{TH_lW_l\times D}$ fixed spatiotemporal positional embeddings, corresponding to every scale level, $l$. We also use learnable scale embeddings upsampled, $\mathsf{P}^{sc}_l \in \mathbb{R}^{1\times D}$, following~\cite{cheng2021masked}. %We repeat the scale embedding at all positions, $T$, $H_l$, $W_l$, resulting in $\hat{\mathsf{P}}^{sc}_l \in \mathbb{R}^{TH_lW_l \times D}$.

%Thus, we perform multiscale processing while maintaining the spatiotemporal dimension for the output, $\mathsf{O}_l \in \mathbb{R}^{TH_lW_l \times D}$, as illustrated in Fig.~\ref{fig:multigrid_multiscalemem} bottom. This mode of operation ensures cross-scale communcation with the detailed feature maps and maintains the spatiotemporal dimension output from our decoder. 
For every scale level, $l$, applying $\mathcal{A}$ will learn to attend among the different set of learnable memory features based on their relevance to the support-query correlation features. It then aggregates the learned read-out memory based on these attention maps to better separate the novel class and the background. Since we want to allow for information exchange across scales, we use a mixing operation,
\begin{equation}
    \bar{\mathsf{X}}_{l^\prime} = \mathcal{M}(\bar{\mathsf{X}}_{l^\prime}, \mathsf{O}_l) = \bar{\mathsf{X}}_{l^\prime} + \mathcal{I}_l^{l^\prime} \mathsf{O}_l,
    \label{eq:int}
\end{equation}
where $\mathcal{I}_l^{l^\prime}$ performs bilinear interpolation to match the size from level $l^\prime$. The cross attention operation, \eqref{eq:st}, is performed consecutively on all $L$ levels and is repeated $N_d$ times, with $N_d$ a hyperparameter denoting the number of decoder layers for each level.

Finally, we define our multiscale comparator, $\mathcal{D}_\text{multiscale}$, as the iterative composition of all cross attention operations, \eqref{eq:st}, and mixing operations, \eqref{eq:int}, to produce the finest scale features, $\mathsf{O}_L$, as
\begin{equation}
\mathsf{O}_L = \mathcal{D}_{\text{multiscale}}(\{\mathsf{\bar{X}}_l\}_{l=1}^L, \mathsf{M}^{\text{memory}}),
\label{eq:comparator}
\end{equation}
which will be used later for the final prediction.
%\textcolor{red}{$\mathsf{O}_L$ is not actually defined until line 523, but it should be here instead. As part of that, need to show explicitly how it is built from the $\mathsf{O}_l$.}

Typically, previous work has focused on multiscale processing in the transformer decoder with multiscale query decoding that contextualizes a set of learnable features, $\mathsf{M}^{\text{query}} \in \mathbb{R}^{N \times D}$  \cite{cheng2021masked,cheng2021mask2former}. In that case, the multiscale query decoding output can be seen as
\begin{equation}
    \mathsf{O}^{\text{query}}_l = \mathcal{A}(\mathsf{M}^{\text{query}} + \mathsf{P}, \bar{\mathsf{X}}_l + \mathsf{P}^{sc}_l + \mathsf{P}^{st}_l, \bar{\mathsf{X}}_l),
    \label{eq:target}
\end{equation}
where $\mathsf{O}^{\text{query}}_l \in \mathbb{R}^{N \times D}$ are a set of compressed learnable queries that are exchanged across scales. This formulation uses learnable features, $\mathsf{M}^{\text{query}}$, as queries and the multiscale feature maps, $\bar{\mathsf{X}}_l$, as keys and values resulting in outputs per scale of dimension, $N\times D$. In contrast, our formulation of multiscale processing with a memory module uses the learnable features, $\mathsf{M}^{\text{memory}}$, as keys and values; hence, the queries are the detailed feature maps, $\bar{\mathsf{X}}_l$, which yields per-scale output of dimension, $TH_lW_l\times D$. These differences are illustrated in Fig.~\ref{fig:multiscalemem}. Thus, the detailed feature maps necessary for segmentation are lost during the communication across scales in the former, but are preserved in ours. We empirically validate this distinction in Sec.~\ref{sec:ablation}. We describe our memory issues and their solutions in the appendix.
%Moreover, we demonstrate solutions to the arising memory problems with our mechanism in the supplement.

\begin{figure}[t]
\vspace{-1em}
    \centering
       \includegraphics[width=0.45\textwidth]{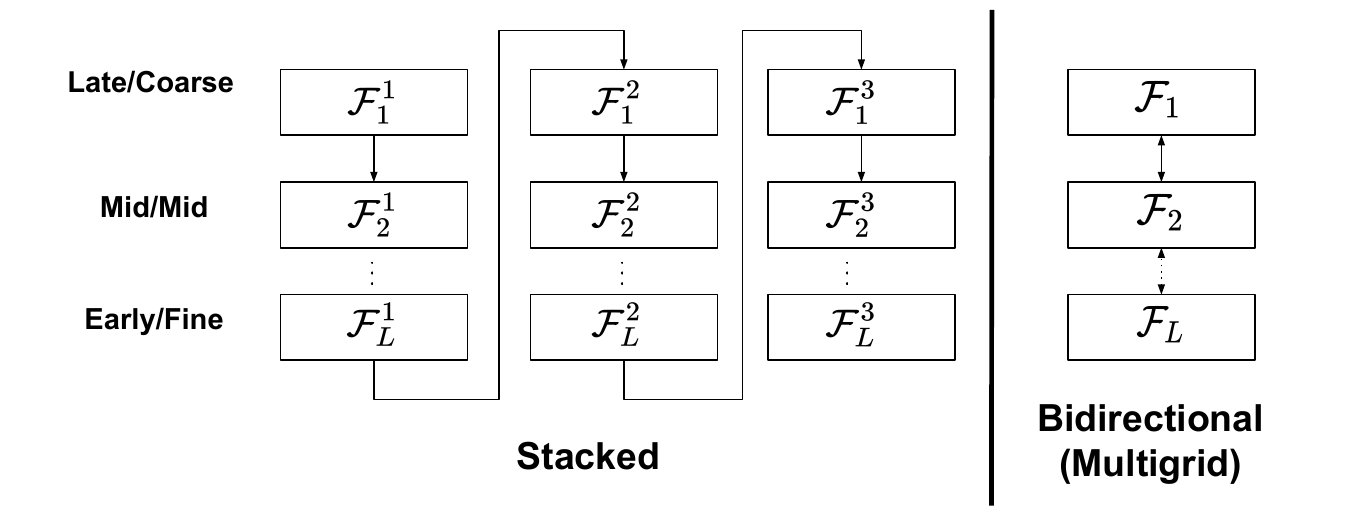} 
    \vspace{-1em}
    \captionsetup{labelfont=bf}
    \caption{Two variants of our information exchange across scales with different feature abstraction/resolution levels: \textit{stacked} \vs \textit{multigrid}. $\mathcal{F}_l$ denotes cross attention, \eqref{eq:st}, for level $l$ followed by mixing, %bilinear interpolation, 
    \eqref{eq:int}. In the \textit{stacked} variant, $\mathcal{F}^j_l$ is for the $j^{th}$ iteration.}
    \label{fig:multigrid}
    \vspace{-1em}
\end{figure}

\subsection{Multiscale multigrid attention (MMA)}\label{sec:multigrid}
We investigate multiple forms of information exchange as variants of our approach and in section~\ref{sec:ablation} we provide empirically-based insights on which information exchange scheme to use in different tasks. Communication across scales can be conducted coarse-to-fine in a \textit{stacked} manner for multiple layers, Fig.~\ref{fig:multigrid} (left), or a \textit{multigrid} manner that allows for bidirectional information transfer between the coarse and fine scales,  Fig.~\ref{fig:multigrid} (right).

Bidirectional exchange ensures that the coarse-grained smoothed correlation features with high level semantics can modulate information in the fine-grained ones, while allowing fine detailed information to affect the coarse-grained.
Inspired by classical multigrid methods \cite{briggs2000multigrid}, we present a multiscale, multigrid attention module that allows for bidirectional information transfer across different scales. In our case, since the multiscale, multigrid attention operates on correlation tensors from different scales, it ensures smooth solutions through bidirectional information exchange between coarse and fine scales. It thereby avoids erroneous correlations that might exist at the different scales, where the fine scale can exhibit erroneous correlations as it only captures low level semantics and the coarse scale can exhibit noisy correlations from being a subsampled signal. 
%\textcolor{blue}{remove this?" Unlike classical multigrid methods our multiscale input exhibits both different levels of resolution as well as different feature abstraction levels (\ie early, mid and late stage features), instead of solely using different sampling rates on the same information as classical multigrid. Another perspective that motivates the coarse-to-fine and fine-to-coarse communication is inspired from earlier work that has shown convolutional layers mainly act as a band pass filtering~\cite{hadji2020convolutional}, which indicates that features from every layer capture a certain set of frequency components. Thus, the bidirectional information transfer can enrich the features through exchanging information across the different set of frequencies captured at each layer.} %Thus, we believe the bidirectional information exchange in our case serves an additional purpose where it allows bidirectional exchange between high-level semantics from late stage features and low-level semantics from early ones.

%\textcolor{green}{Seems overly redundant to once again refer to Fig 3(a) vs 3(b) again, when they were compared just one paragrap before.}
%In Fig.~\ref{fig:multigrid} we show our proposed \textit{multigrid} bidirectional information flow \vs simple \textit{stacked} coarse-to-fine processing. 
To be more precise, consider \textit{stacked} multihead attention, which takes as input multiscale pyramids. It is composed of a series of multihead cross attentions, \eqref{eq:st}, followed by merging the two consecutive levels and using bilinear interpolation to match the scale, \eqref{eq:int}. The combined aforementioned operations can be denoted as, $\mathcal{F} = \mathcal{M} \circ \mathcal{A}$, where $\circ$ denotes function composition and is applied on the features from scale level, $l \in \{1,2, \dots, L\}$ as, $\mathcal{F}_l$. Thus, we formulate \textit{stacked} multihead attention (SMA) in a coarse-to-fine processing per iteration as %Eq.~\ref{eq:sma}, \vs multigrid, Eq.~\ref{eq:mma}, at 
\begin{equation}
    \text{SMA} = \mathcal{F}^j_L \circ \dots \mathcal{F}^j_2 \circ \mathcal{F}^j_1,
    \label{eq:sma}
\end{equation}
where $\mathcal{F}^j_l$ corresponds to the $j^{th}$ iteration as the operations in the \textit{stacked} multihead attention are repeated $N_d$ times. In comparison, 
%On the other hand, the 
\textit{multigrid} multihead attention (MMA) processing is defined as
\begin{equation}
\text{MMA} = \mathcal{F}_L \circ \mathcal{F}_{L-1} \circ \dots \circ \mathcal{F}_2 \circ \mathcal{F}_1 \circ \mathcal{F}_2  \circ \dots \circ \mathcal{F}_L \circ \dots \circ \mathcal{F}_2 \circ \mathcal{F}_1.
\label{eq:mma}
\end{equation}
Thus, the multigrid approach, \eqref{eq:mma}, encompasses bidirectional information exchange across scales, while the stacked approach is strictly coarse-to-fine.
%It is seen that the multigrid approach, \eqref{eq:mma}, encompasses bidirectional information exchange across scales similar to classical recursive methods~\cite{briggs2000multigrid}, whereas the stacked approach is strictly coarse-to-fine. 
%statcThis \textit{multigrid} formulation is the final design we use in our multiscale comparator, hence the name multiscale multigrid comparator. 
Sec.~\ref{sec:ablation} provides empirical support for the robustness of the \textit{multigrid} formulation. % proves to be superior to the \textit{stacked} one emperically.

\vspace{5pt}
\subsection{Boundary refinement}
\label{sec:bdry_refine}
The finest scale feature output, $\mathsf{O}_L$, is obtained from correlation tensors between the support and target query sets. While this formulation facilitates comparisons between the support and query, it can loose detailed information in the original backbone features that would facilitate delineation of precise object boundaries.
%The finest scale features output after being enhanced in the multiscale memory learning and multigrid information exchange, $\mathsf{O}_L$, goes through a transformer based refinement operation. Since, $\mathsf{O}_L$, is obtained from correlation tensors between the support and target query sets it can lose some detailed information necessary to segment the objects boundaries. 
In response, we augment $\mathsf{O}_L$ with the target query video, $\mathsf{I}^{(q)}$, and employ a refinement phase using a transformer architecture, $\mathcal{D}_{\text{refine}}$, to improve the predicted segmentation, $\hat{\mathsf{M}}$, and its boundary, $\partial\hat{\mathsf{M}}$, 
as 
\begin{equation}
<\hat{\mathsf{M}}, \partial\hat{\mathsf{M}}> \hspace{5pt} = \mathcal{D}_{\text{refine}}(\mathsf{I}^{(q)}, \mathsf{O}_L),
\label{eq:bdry}
\end{equation}
Here, $\mathcal{D}_{\text{refine}}$ entails multiple attention blocks, with self supervised pretraining~\cite{caron2021emerging}, followed by two parallel convolutional prediction heads to produce the segmentation mask and its boundary as an ordered pair $<\hat{\mathsf{M}}, \partial\hat{\mathsf{M}}>$. %Details are provided in the supplement.
%The refined features go through two prediction heads, $\mathcal{D}_{\text{seg}}, \mathcal{D}_{\text{bdry}}$, guided with the comparator output, $\mathsf{O}_L$, to simultaneously predict the segmentation mask and the boundary, $\hat{\mathsf{M}}, \hat{\mathsf{M}}_{\text{bdry}}$, resp. as follows,

%\begin{subequations}
%\begin{equation}
%\hat{\mathsf{M}} = \mathcal{D}_{\text{seg}}(\mathsf{O}_L, \mathsf{R}),
%\end{equation}
%\begin{equation}
%\hat{\mathsf{M}}_{\text{bdry}} = \mathcal{D}_{\text{bdry}}(\mathsf{O}_L, \mathsf{R}),
%\end{equation}
%\end{subequations}

\section{Learning scheme}
In this section, we start by summarizing the meta-learning setup and our architecture overview for FS-VOS. Subsequently, we describe its extension to AVOS.

%\subsection{Few-shot setup}
%\textcolor{green}{In the interest of time, I'm going to assume that this subsection is okay as is, because presumably its based on our previous writing explaining the set-up.}\textcolor{blue}{menna:Yes it is, I highlighted the changed part in blue to accomodate both tasks}
\subsection{Meta-learning setup} 
\label{sec:metalearning_setup}
We formulate the FS-VOS task as follows~\cite{chen2021delving,yang2021few}. 
%Let $D_{\text{train}}$ and $D_{\text{test}}$ be training and testing data, resp. 
For every dataset, we split the $C$ categories into $O$ folds, each fold will have $\frac{C}{O}$ novel categories, $\mathcal{C}_{\text{test}}$, and $C - \frac{C}{O}$ as base classes, $\mathcal{C}_{\text{train}}$. Both the training and test classes do not intersect. Meta-learning consists of two phases: meta-training and meta-testing. In the meta-training phase, we sample $N_{e}$ tasks from the corresponding dataset with support and target query set pairs $\{\mathcal{S}_i, \mathcal{Q}_i\}_{i=1}^{N_{e}}$ for classes in $\mathcal{C}_{train}$. Similarly in meta-testing we sample these paired sets but for classes in $\mathcal{C}_{test}$. The target query set contains video frames $\mathcal{Q}=\{\mathsf{I}^{(q)}\}$, where superscript $\_^{(q)}$ denotes the target query set, $\mathsf{I}^{(q)}\ \in \mathbb{R}^{T \times H \times W \times 3}$. The support set in a one-way $K$-shot task has $K$ image-label pairs $\mathcal{S}=\{\mathsf{I}^{(s)}_k, {\mathsf{M}_k}^{(s)}\}_{k=1}^K$. The superscript $\textbf{\_}^{(s)}$ denotes support set and ${\mathsf{M}_k}^{(s)}$ is a binary segmentation mask for the class considered. The image-label pairs $\mathsf{I}_k \in \mathbb{R}^{H \times W \times 3}$ and $\mathsf{M}_k \in \mathbb{R}^{H \times W}$, with $H\times W$ spatial dimensions.
%In the case of video actor/action segmentation the one-way $K$-shot task has $K$ trimmed video-label pairs $\mathcal{S}=\{\mathsf{I}^{(s)}_k, {\mathsf{M}_k}^{(s)}\}_{k=1}^K$. The binary segmentation mask ${\mathsf{M}_k}^{(s)}$ is for one frame in the trimmed video for an actor/action class. Thus, the video-label pairs are $\mathsf{I}_k \in \mathbb{R}^{T \times H \times W \times 3}$ and $\mathsf{M}_k \in \mathbb{R}^{H \times W}$.

%\subsection{Meta-learning a multiscale comparator}
\subsection{Meta-learning a multiscale video comparator}
\label{sec:method_overview}

%\textcolor{red}{RW: Somewhere in this subsection the term \textit{meta-learning} must appear explicitly with explanation of how it applies to the rest of the subsection, \eg exactly where and how is meta-learning applied? It cannot be assumed that its implicitly obvious.}
We begin with a review of the full architecture of our multiscale comparator and the learning process, as shown in Fig.~\ref{fig:detailed}. We initially assume a one-shot setting and subsequently discuss the K-shot extension. We use a pretrained convolutional backbone with fixed weights, $\mathcal{B}$, that are not updated during the meta-training process, to compute the support and target query set features. The feature pyramid for the one-shot support and target query sets are extracted from multiple layers, as, $\mathsf{F}^{(s)} = \mathcal{B}(\mathsf{I}^{(s)}), \mathsf{F}^{(q)} = \mathcal{B}(\mathsf{I}^{(q)})$, resp. We mask the support set feature pyramid with the groundtruth mask, $\mathsf{M}^{(s)}$, resulting in a masked support feature pyramid, $\hat{\mathsf{F}}^{(s)}$. Let $\phi(\cdot,\cdot)$ denote the (hyper)correlation encompassing the comparisons between its arguments, both of which are pyramids of tensors. Then, we define a 4D hypercorrelation tensor pyramid with $L$ levels as $\{\mathsf{H}_l\}_{l=1}^L = \phi(\hat{\mathsf{F}}^{(s)}, \mathsf{F}^{(q)})$, \cf~\cite{min2021hypercorrelation}. We use the hypercorrelation squeeze network~\cite{min2021hypercorrelation}, $\mathcal{D}_{\text{hypercorr}}$ that provides efficient 4D convolution and generates a 2D feature pyramid, which subsequently is flattened further, $\{\mathsf{\bar{X}}_l\}_{l=1}^{L} = \mathcal{D}_{\text{hypercorr}}( \{\mathsf{H}_l\}_{l=1}^{L} )$. During training, we sample support and query sets to emulate the inference stage (\ie we employ meta-training).

%Let $$\phi(.,.)$ coperation denotes computing the correlation tensors and $\bigoplus$ is the concatenation on the channel dimension for $m$ consecutive layers that have the same spatial dimensions.In the hypercorrelation pyramid, each level is the concatenation of the correlation tensors from different layers, $H_p = \bigoplus_l^{l+m} \phi(f_l^{(s)}, f_l^{(q)})$. The $\phi$ operation denotes computing the correlation tensors and $\bigoplus$ is the concatenation on the channel dimension for $m$ consecutive layers that have the same spatial dimensions. We use the hypercorrelation squeeze network~\cite{min2021hypercorrelation} as our neck, similar to the feature pyramid network~\cite{lin2017feature} in fully supervised models, to perform 4D convolution and generate a feature pyramid, $\{Z_p\}_{p=1}^{P} = D_{\text{hypercorr}}( \{H_p\}_{p=1}^{P} )$, based on the hypercorrelation pyramid. %The constructed feature pyramid is used as input to our multiscale transformer decoder. 

Our multiscale comparator transformer uses the features extracted on different levels by performing cross attention to a learnable memory, $\mathsf{M}^{\text{memory}}$, to yield the final output feature maps, $\mathsf{O}_L$, ~\eqref{eq:comparator}. %= \mathcal{D}_{\text{multiscale}}(\{\mathsf{\bar{X}}_l\}_{l=1}^L, \mathsf{M}^{\text{memory}})$. 
%\textcolor{green}{So, $O_P =\{Z_p\}_{p=1}^{P}$, which is the dictionary? If so, then don't use multiple terms for the same thing.}\textcolor{blue}{menna: So B is the memory and $O_P$ is the final output.} 
The cross attention reweighs the memory features to enhance the query feature maps and separate the background from the novel class based on their correlation to the support set. Attention and feature aggregation subsequently is computed in a pixel-wise manner across all levels to produce the final output features, $\mathsf{O}_L$. 
%\textcolor{green}{Again, very confusing, you defined $O_P$ at the start of the paragraph, which I think actually was defined at the end of the previous paragraph, but now you seem to be adding more or even changing the explanation.}\textcolor{blue}{menna:let me know if after the modification I did above if this is still confusing.} 

\begin{table*}[t!]
\centering
\small
\begin{tabu}{@{}lccccccccccc@{}}
\tabucline[1pt]{-}
\multirow{2}{*}{Method} & \multicolumn{5}{c}{$\mathcal{J}$} & \multicolumn{5}{c}{$\mathcal{F}$} & \multirow{2}{*}{$\mathcal{F} \& \mathcal{J}$} \\ \cmidrule(r{2pt}){2-6}  \cmidrule(r{2pt}){7-11}
&  1 & 2 & 3 & 4 & Mean & 1 & 2 & 3 & 4 & Mean\\ \tabucline[1pt]{-}
PMMs~\cite{yang2020prototype}& 32.9 & 61.1 & 56.8 & 55.9 & 51.7 & 34.2 &  56.2 & 49.4 & 51.6 & 47.9 & 49.8\\
PFENet~\cite{tian2020prior} & 37.8 & 64.4 & 56.3 & 56.4 & 53.7 & 33.7 &  55.9 & 48.7 & 48.9 & 46.8 & 50.3\\
PPNet~\cite{liu2020part} & 45.5 & 63.8 & 60.4 & 58.9 & 57.1& 35.9 &  50.7&  47.2& 48.4 & 45.6 & 51.4 \\
%DANet w/o OL~\cite{chen2021delving} & 41.5 & 64.8 & 61.3 & 61.4 & 57.2 & 40.3 & 62.3 & 60.2&  59.4&  55.6 & 56.4\\
DANet~\cite{chen2021delving} & 43.2 & 65.0 & 62.0 & 61.8 & 58.0 & 42.3 & \textbf{62.6} &  \textbf{60.6} & 60.0 & 56.3 & 57.2\\\hline
%MMC transformer (ours) & \textbf{52.8} & \textbf{72.4} & \textbf{64.1} & \textbf{66.1} & \textbf{63.9} & \textbf{45.2} & 60.2 & 53.5 & 57.8 & 54.2 & \textbf{59.1}\\ \tabucline[1pt]{-}
\textbf{MMCT (ours)} & \textbf{52.2} & \textbf{73.8} & \textbf{65.7} & \textbf{68.7} & \textbf{65.1} & \textbf{45.5} & 62.1 & 57.8 &	\textbf{60.3} & \textbf{56.4} & \textbf{60.8}\\ \tabucline[1pt]{-}
\end{tabu}
\captionsetup{labelfont=bf}
\captionof{table}{Comparative FS-VOS results on YouTube-VIS over four folds with a five-shot support set as mIoU, $\mathcal{J}$, and mean boundary accuracy, $\mathcal{F}$, and their average, $\mathcal{F} \& \mathcal{J}$. MMCT: MMC Transformer.} %\textcolor{red}{What is $\mathcal{F} \& \mathcal{J}$?}}
%\vspace{-1.3em}
\label{table:soa_fsvos}
\vspace{-1em}
\end{table*}

Our multiscale memory transformer decoder enriches the features and allows bidirectional exchange of information across scales. The output features, $\mathsf{O}_L$, from the multiscale comparator are used in a boundary refinement module, $\mathcal{D}_{\text{refine}}$, along with the target query video, $\mathsf{I}^{(q)}$, which results in the final segmentation and boundary predictions, $\hat{\mathsf{M}}, \hat{\mathsf{M}}_{\text{bdry}}$, resp.
The predictions, $\hat{\mathsf{M}}, \hat{\mathsf{M}_{\text{bdry}}}$, are for the novel and background class. The hypercorrelation squeeze network, $\mathcal{D}_{\text{hypercorr}}$, our multiscale comparator, $\mathcal{D}_{\text{multiscale}}$, and the boundary refinement, $\mathcal{D}_{\text{refine}}$, are meta-trained with a simple binary cross entropy and boundary loss. For the boundary loss, the groundtruth segmentation mask is processed with morphological operations to provide the groundtruth boundary. During few-shot inference when operating with K-shot support set, we re-weigh the predictions from the $K$ examples and divide by the weights summation, which we call an adaptive K-shot scheme. This scheme alleviates challenges arising from confusing support examples by downweighting their influence. The weight is assigned based on the similarity between the pooled coarsest support features, $\mathsf{F}^{(s)}_1$, with the groundtruth, $\mathsf{M}^{(s)}$, and target query features, $\mathsf{F}^{(q)}_1$, using the prediction, $\hat{\mathsf{M}}$, as a soft mask. In particular, the weight of each support example, $k$, is
\begin{equation}
w_k = \mathcal{C}(\mathcal{P}(\mathsf{F}^{(s)}_1, \mathsf{M}^{(s)}), \mathcal{P}(\mathsf{F}^{(q)}_1, \hat{\mathsf{M}})),
\label{eq:adaptive_kshot}
\end{equation}
where $\mathcal{P}$, is  masked pooling and $\mathcal{C}$ is cosine similarity. %The weight is assigned based on the similarity between the pooled coarsest support features, $\mathsf{F}^{(s)}_1$, with the groundtruth, $\mathsf{M}^{(s)}$, and target query features, $\mathsf{F}^{(q)}_1$, using the prediction, $\hat{\mathsf{M}}$, as a soft mask.%\textcolor{red}{This reweighting scheme needs its own numbered equation, especially because you come back to it in the ablations.}

\textbf{Extensions to additional tasks.} The modifications required to apply our approach to AVOS
%fewshot actor/action segmentation and , 
occur in the construction of the input tensors to the MMC Transformer. 
%For the case of fewshot actor/action segmentation, support-query correlations tensors between two backbone features are used. The first backbone captures spatial features from the centered image in the clip, while the second backbone captures spatiotemporal features from the entire clip. The correlation tensors from both backbones are concatenated on the channel dimension. We find this change necessary because, unlike semantic segmentation, action segmentation depends not only on spatial features, but also spatiotemporal features (\eg motion). 
%For the fully supervised AVOS and actor/action segmentation, 
We simply use the feature pyramid from the backbone as direct input to our multiscale transformer decoder. This change is necessary because, unlike few-shot inference, we do not have few labelled class exemplars and therefore must base predictions directly on the backbone features. %During inference, in actor/action segmentation we apply a sliding window and predict for the centre frame. 
%Extensions to actor/action segmentation are provided in the supplement.

\section{Empirical results}
\subsection{Experiment design}

\textbf{Datasets.} We evaluate on YouTube-VIS FS-VOS~\cite{chen2021delving} to facilitate comparison to FS-VOS the state of the art. We follow standard evaluation protocols, with details in the appendix. We present AVOS results on DAVIS'16~\cite{Perazzi2016}, MoCA~\cite{lamdouar2020betrayed} and YouTube-Objects (YTBO)~\cite{prest2012learning}. The challenging MoCA dataset tests the ability of models to capture dynamics and segment camouflaged moving animals. %Actor/action segmentation A2D~\cite{xu2015can,yang2021few} results are saved for the supplement. 
%\textcolor{red}{Even in the supplement,  all results presented must be comparable to the comparison approaches.}

\textbf{Implementation details.}
\label{sec:detailimp}
For FS-VOS we follow the same architectural choices as state-of-the-art approaches~\cite{chen2021delving,liu2020part,tian2020prior,yang2020prototype} to facilitate comparison, where we build on a ResNet-50~\cite{HeZRS16} backbone pretrained on ImageNet~\cite{deng2009imagenet}. We also build a conventional multiscale baseline for FS-VOS with coarse-to-fine convolutional processing using previous work on few shot image segmentation~\cite{min2021hypercorrelation} that we adapt to video by meta-training on sampled support images and query videos, which we call \textit{Multiscale Baseline}.
%For few-shot actor/action segmentation, we use both ResNet-50 pretrained on ImageNet~\cite{deng2009imagenet} and X3D~\cite{feichtenhofer2020x3d} pretrained on Kinetics~\cite{kay2017kinetics} for the static and dynamic backbones, resp. 
In our MMC Transformer, the only hyperparameters are the number of decoder layers, $N_d = 3$, and memory entries, $N = 20$. We meta-train our model and multiscale baseline % with 50 epochs 
on YouTube-VIS.
%, while we use 70 epochs for A2D. 
We use the same hyperparameters for both our approach and the baseline. %where we use AdamW~\cite{loshchilov2018decoupled} with a learning rate of $1\times10^{-3}$ and weight decay of $1 \times 10^{-4}$. 
For AVOS we build a baseline that takes an input clip of RGB images, extracts features from Video-Swin~\cite{liu2022video} followed by pixel decoding~\cite{lin2017feature} and the previously proposed multiscale transformer decoding~\cite{cheng2021masked}, which we call \textit{Multiscale Baseline (MQuery-Stacked)}. The appendix has more implementation details. % \textcolor{red}{Similar to or exactly that. Also, need to justify use of different back bones for the two tasks.}

\subsection{Comparison to state-of-the-art approaches}
\textbf{Few-shot video object segmentation.} Table~\ref{table:soa_fsvos} provides a comparison with existing approaches on YouTube-VIS FS-VOS. Our approach shows a notable gain of 7.1\% with respect to the recently presented single scale many-to-many attention comparator (DANet~\cite{chen2021delving}) in terms of mean intersection over union (mIoU), while maintaining on-par performance on the mean boundary accuracy. This result demonstrates that our multiscale memory comparator %which meta-learns a memory module and enahnces the query features at multiple levels of resolution and feature abstraction can 
helps separate the novel class with respect to the background better than previous approaches. On Fold 3 our boundary accuracy is worse than DANet, due to challenging scenarios where the objects undergo occlusion or motion blur. Nevertheless, our overall mean boundary accuracy remains on-par to DANet. Moreover, we show our approach outperforms DANet in inference time in the appendix.

%\textbf{Actor/action segmentation.}
%Table~\ref{table:ablation_a2d_fsvos} shows comparisons with the state of the art for few-shot video actor/action segmentation~\cite{yang2021few,siam2020weakly}. It is seen that our approach consistently outperforms the others in both the one-shot and five-shot scenarios on Common A2D. However, note that our method and baselines use an additional information with a groundtruth segmentation of one frame in the support video, unlike~\cite{yang2021few,siam2020weakly}. \textcolor{blue}{Nonetheless, our main results are on the video segmentation task and we only use actor/action segmentation to show the flexibility of our approach.} \textcolor{blue}{I am suggesting that we leave out the text comparing to the SOA but leave the table as is and discuss on it the comparison to the baseline as a way to show that our baseline is competitive.}

%Additionally, annotating one frame in a support video requires minimal effort as we do not require the full support video to be annotated. We leave it for future work to investigate weak supervision without the need for pixel-level labels.}\textcolor{red}{We need to discuss this matter. As I understand what is written, it seems like you are saying we get to use additional information, but it doesn't matter -- not a good argument!}

\textbf{Fully supervised AVOS.} To demonstrate the versatility of our multiscale transformer decoder we go beyond few-shot VOS and evaluate on fully supervised AVOS. Table~\ref{table:soa_avos} shows that our approach outperforms the state of the art on two datasets, YouTube Objects and MoCA (the most challenging due to its entailing camouflaged moving animals), with a margin of up to 16\%. Unlike state-of-the-art alternatives, we do not explicitly compute optical flow, as our model learns end-to-end to segment the moving objects from an input clip. For DAVIS when comparing without post processing, our results are only lower than RTNet, which uses an additional saliency segmentation dataset DUTS~\cite{wang2017learning} for pretraining. We leave it for future work to investigate such pretraining. As it stands, however, when we add multiscale inference postprocessing (documented in the supplement) ours already outpeforms RTNet, without using optical flow input or this extra data.
%However, our approach with multiscale inference as a post processing and without using the extra saliency segmentation dataset outperforms RTNet. 
AVOS results with additional metrics are provided in the supplement.
%Additionally, it was shown in~\cite{kowal2022deeper} that DAVIS dataset has more bias towards capturing information from single frame (\ie static bias), thus it is less interesting than MoCA that evaluates the models' ability to capture information conveyed from multiple frames and going beyond appearance. Additional results are provided in the supplementary.

\begin{table}[t!]
\centering
\small
\begin{tabu}{@{}lccc@{}}
\tabucline[1pt]{-}
\multirow{2}{*}{Method} & \multicolumn{3}{c}{mIoU}\\ \cmidrule(r{2pt}){2-4}
& DAVIS~\cite{Perazzi2016} & MoCA~\cite{lamdouar2020betrayed} & YTBO~\cite{prest2012learning} \\ \tabucline[1pt]{-}
AGS~\cite{wang2019learning} & 79.7/- & - & 69.7\\
COSNet~\cite{lu2019see} & 80.5/- & 50.7 & 70.5 \\
AGNN~\cite{wang2019zero} & 80.7/78.9 & - & 70.8\\
MATNet$\dagger$~\cite{zhou2020motion} & 82.4/- & 64.2 & 69.0 \\
%FSNET~\cite{ji2021full} & 83.4/ & - & -\\
%TransportNet~\cite{zhang2021deep} & 84.5/ & & \\
%AMCNet$\dagger$~\cite{yang2021learning} & 84.5/83.0 & - & 71.1\\
RTNet$\dagger$~\cite{ren2021reciprocal} & 85.6/84.3 & 60.7 & 71.0\\ \hline
\textbf{MMCT (ours)} & \textbf{86.1}/83.8 & \textbf{80.3} & \textbf{78.2}\\
%MQuery-Stacked & 83.7/81.3 & 77.4 & 76.8 \\ \hline
%\textbf{MMemory-Stacked} & \textcolor{red}{86.1}/83.5 & \textcolor{red}{80.3} & \textcolor{red}{78.2}\\
%\textbf{MMemory-Multigrid} & 85.4/\textcolor{blue}{83.7} &  \textcolor{blue}{79.3} & \textcolor{blue}{78.1}\\
%\textbf{MMemory-Multigrid}$\ddagger$ & \textcolor{blue}{85.1} &  \textcolor{blue}{79.7} & \color{red}{78.2}\\
%Ours R101 & 81.8	& 66.6 & 73.5\\ 
%Ours$\dagger$ & \textbf{83.5} & \textbf{80.3} & \textbf{78.2}\\
\tabucline[1pt]{-}
\end{tabu}
\captionsetup{labelfont=bf}
\captionof{table}{Comparison to the state of the art on the AVOS task with three datasets reporting mean intersection over union. $\dagger$ indicates methods using optical flow. On DAVIS we report results with/without postprocessing. MMCT: MMC Transformer.}
%MQuery is multiscale query decoding, MMemory is multiscale memory decoding. Our two variants of multiscale memory transformer decoding are bolded. 
%Best and second best results are highlighted in red and blue resp.}
%\textcolor{red}{Its confusing to combine the ablation and SOA comparison: Use separate tables.}} 
%using ResNet-101 backbone. $\dagger$ indicates our model with Video-Swin backbone as the best reported results.}
\label{table:soa_avos}
\vspace{-1.2em}
\captionsetup{labelformat=empty}
\end{table}

\subsection{Ablation study}
\label{sec:ablation}

%We use ablations to document the distinct benefits of our multiscale memory decoding, information exchange across scales, multiscale comparators and boundary refinement. For the first two components we use the densely annotated per-frame datasets, YouTube-VIS FSVOS and AVOS datasets \textcolor{blue}{because their dense annotations can evaluate better the effectiveness of these components in dense prediction tasks}. For the third component, we use both few-shot, tasks, FSVOS and action/actor segmentation, as they investigate our approach within the intended meta-learning framework. \textcolor{blue}{For the last component, we focus on the FSVOS, because boundary refinement is necessary when our input is correlation tensors and because the state of the art in FSVOS is concerned with the boundary accuracy.} %\textcolor{red}{Last point unclear.}

Our comparison to the state of the art documented the overall strength of MMC Transformer, which confirms our first contribution: state-of-the-art performance using our multiscale comparator for few-shot video tasks. In this section, we use ablations to document the distinct benefits of our multiscale memory decoding, information exchange across scales, multiscale comparator and boundary refinement. We ablate on both AVOS and FS-VOS. 
%For multiscale memory decoding and information across scales, we also ablate with respect to fully supervised AVOS, as application of our model to that task illustrates the benefits of those innovations. 
For AVOS we only ablate multiscale memory decoding and information across scales, since the other components are not relevant to the fully supervised task: There is no support set to enable the comparator; boundary refinement is defined to combat the blurring entailed in use of correlation tensors~\cite{hong2022cost}.
%rather than backbone features.})
%the video object segmentation tasks the FS-VOS and the fully supervised AVOS. However, the multiscale comparator, the adaptive k-shot scheme and the boundary refinement are only ablated in the FS-VOS task since they are either not applicable or not needed in the AVOS. The multiscale comparator and adaptive k-shot scheme are tightly coupled with the few-shot problem within a meta-learning framework. While the boundary refinement is only necessary because of our use of support-query correlation tensors as input in the FS-VOS. It was also observed in concurrent work~\cite{hong2022cost} that correlation tensors do not capture detailed boundaries and require the original query features to refine it further.}
%For the first two components we use the densely annotated per-frame datasets, YouTube-VIS FSVOS and AVOS datasets because their dense annotations can evaluate better the effectiveness of these components in dense prediction tasks. For the third component, we use both few-shot, tasks, FSVOS and action/actor segmentation, as they investigate our approach within the intended meta-learning framework. For the last component, we focus on the FSVOS, because boundary refinement is necessary when our input is correlation tensors and because the state of the art in FSVOS is concerned with the boundary accuracy.}

\begin{table}[t!]
\centering
\small
\begin{tabu}{@{}lccc@{}}
\tabucline[1pt]{-}
\multirow{2}{*}{Method} & \multicolumn{3}{c}{mIoU}\\ \cmidrule(r{2pt}){2-4}
& DAVIS~\cite{Perazzi2016} & MoCA~\cite{lamdouar2020betrayed} & YTBO~\cite{prest2012learning} \\ \tabucline[1pt]{-}
\makecell[lt]{Multiscale Baseline\\ (MQuery-Stacked)} & \makecell[c]{\\83.7/81.3} &  \makecell[c]{\\77.4} &  \makecell[c]{\\76.8} \\ \hline
\textbf{MMemory-Stacked} & \textbf{86.1}/83.8 & \textbf{80.3} & \textbf{78.2}\\
\textbf{MMemory-Multigrid} & 85.4/\textbf{84.0} &  79.3 & 78.1\\
%\textbf{MMemory-Multigrid}$\ddagger$ & \textcolor{blue}{85.1} &  \textcolor{blue}{79.7} & \color{red}{78.2}\\
%Ours R101 & 81.8	& 66.6 & 73.5\\ 
%Ours$\dagger$ & \textbf{83.5} & \textbf{80.3} & \textbf{78.2}\\
\tabucline[1pt]{-}
\end{tabu}
\captionsetup{labelfont=bf}
\captionof{table}{Ablation study on the AVOS task with three datasets reporting mean intersection over union. MQuery, MMemory denote multiscale query and memory decoding resp. Our variants are bolded.} 
%using ResNet-101 backbone. $\dagger$ indicates our model with Video-Swin backbone as the best reported results.}
\label{table:ablation_avos}
\vspace{-1.2em}
\captionsetup{labelformat=empty}
\end{table}

\textbf{Multiscale memory decoding.} 
%\textcolor{red}{I still find this entire ablation confusing. I suggest starting with the FS-VOS case, because its the primary focus. In the text, walk through the relevant rows in the table, row-by-row. After establishing the benefits there, then use AVOS to further the case. Also, still need to be clearer about what are the baselines? If its previous work, then need to explain clearly that our model reduces to those under the ablation and they need to be cited in the tables.}
We first consider our main contribution, multiscale memory decoding, which preserves the spatiotemporal dimension during multiscale processing, \eqref{eq:st}, compared to multiscale query decoding, \eqref{eq:target}.
%\textcolor{blue}{To build our baselines in the transformer decoding scheme we use the previously proposed multiscale query decoding~\cite{cheng2021masked} (\textit{MQuery-Stacked}) as in, \eqref{eq:target}, which was shown to outperform the single scale decoding.} 
%\textcolor{red}{Still confusing: You already defined baselines in implementation details, but now you are doing it again. Also, need to be clear whether or not you literally use ~\cite{cheng2021masked} and if yes, then add its reference to citation to the tables.}
%\textcolor{red}{Confusing: Is that the baseline that you referred to in the previous paragraph? If not, then where is the baseline? If so, then be say so.} 

Table~\ref{table:ablation_avos} shows results for the fully supervised AVOS task where, our \textit{Multiscale Baseline (MQuery-Stacked)} already outperforms the state of the art on two datasets by a nontrivial margin, \cf Table~\ref{table:soa_avos}. (Note that we do not further ablate \textit{MQuery-Stacked} by removing the multiscale query decoding to reduce to an even simpler multiscale model, since it was shown previously that the addition of multiscale query decoding was beneficial for fully-supervised VOS~\cite{cheng2021masked,cheng2021mask2former}.)
%\textcolor{red}{The baseline introduced in the previous paragraph? If yes, then name it back there, not here. If no, then its confusing to define two baselines for the same AVOS task.} 
More importantly, our multiscale memory decoding \textit{MMemory-Stacked} consistently outperforms the strong baseline \textit{Multiscale Baseline (MQuery-Stacked)} across the three datasets with up to 2.9\% gain. 

\begin{table}[t!]
\centering
\small
\begin{tabu}{@{}lccccc@{}}
\tabucline[1pt]{-}
\multirow{2}{*}{Method} & \multicolumn{5}{c}{mIoU}\\ \cmidrule(r{2pt}){2-6}
& 1 & 2 & 3 & 4 & Mean \\ \tabucline[1pt]{-}
Multiscale Baseline & 49.5 & 69.5 & 63.8 & 65.2 & 62.0 \\
$+$ MQuery-Stacked & 49.4 & 70.5 & 62.9 & 64.3 & 61.8\\ \hline
$+$ \textbf{MMemory-Stacked} & 50.7 & 69.8 & 63.2 & 64.4 & 62.0 \\
$+$ \textbf{MMemory-Multigrid} & 51.5 & 70.6 & 63.0 & 64.6 & 62.4\\
$+$ \textbf{MMemory-Multigrid$\dagger$} & \textbf{52.8} & 72.4 & 64.1 & 66.1 & 63.9\\ 
$+$ \textbf{MMemory-Multigrid$\ddagger$} & 52.2 & \textbf{73.8} & \textbf{65.7} & \textbf{68.7} & \textbf{65.1} \\ \tabucline[1pt]{-}
\end{tabu}
\captionsetup{labelfont=bf}
\captionof{table}{Ablation study on YouTube-VIS FS-VOS four folds with a five shot support set. MQuery is multiscale query decoding, MMemory is multiscale memory decoding. $\dagger$ indicates the adaptive K-shot selection, $\ddagger$ indicates an additional boundary refinement. 
%5\textcolor{red}{Not a clean study: All you want to document is the benefit of multiscale memory decoding, the rest is noise or else would require much elaboration of each and every variant in the text. See also my comments in the text Re. this study.} 
}
\label{table:ablation_decoder_fsvos}
\captionsetup{labelformat=empty}
\vspace{-1em}
\end{table}

\begin{figure*}[t!]
\centering
\noindent
\begin{subfigure}{0.14\textwidth}
    \includegraphics[width=\textwidth]{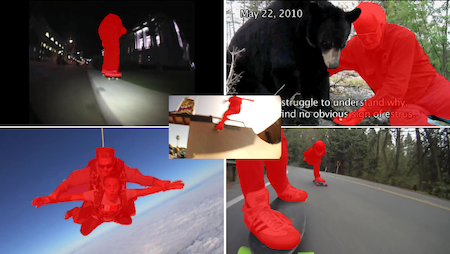}
\end{subfigure}
\begin{subfigure}{0.42\textwidth}
\begin{subfigure}{0.33\textwidth}
    \includegraphics[width=\textwidth]{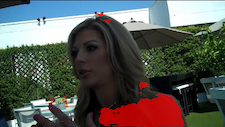}
\end{subfigure}%
\begin{subfigure}{0.33\textwidth}
    \includegraphics[width=\textwidth]{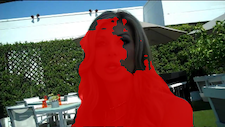}
\end{subfigure}%
\begin{subfigure}{0.33\textwidth}
    \includegraphics[width=\textwidth]{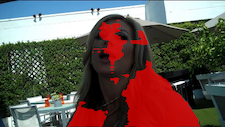}
\end{subfigure}%
\end{subfigure}
\begin{subfigure}{0.42\textwidth}
\begin{subfigure}{0.33\textwidth}
    \includegraphics[width=\textwidth]{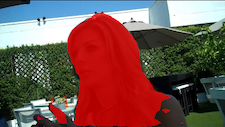}
\end{subfigure}%
\begin{subfigure}{0.33\textwidth}
    \includegraphics[width=\textwidth]{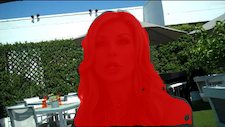}
\end{subfigure}%
\begin{subfigure}{0.33\textwidth}
    \includegraphics[width=\textwidth]{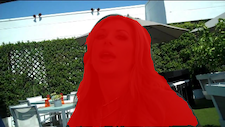}
\end{subfigure}
\end{subfigure}

\begin{subfigure}{0.14\textwidth}
    \includegraphics[width=\textwidth]{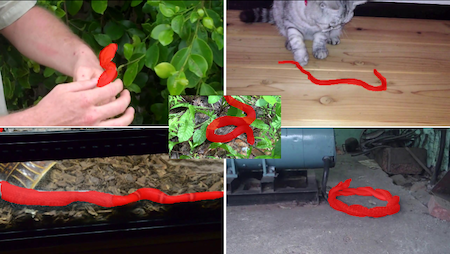}
    \caption{}
\end{subfigure}
\begin{subfigure}{0.42\textwidth}
\begin{subfigure}{0.33\textwidth}
    \includegraphics[width=\textwidth]{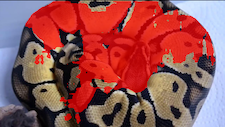}
\end{subfigure}%
\begin{subfigure}{0.33\textwidth}
    \includegraphics[width=\textwidth]{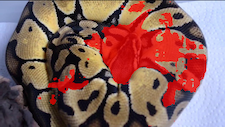}
\end{subfigure}%
\begin{subfigure}{0.33\textwidth}
    \includegraphics[width=\textwidth]{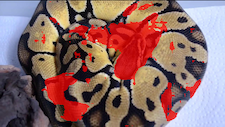}
\end{subfigure}
\caption{}
\end{subfigure}
\begin{subfigure}{0.42\textwidth}%
\begin{subfigure}{0.33\textwidth}
    \includegraphics[width=\textwidth]{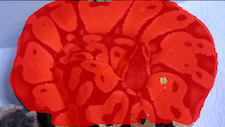}
\end{subfigure}%
\begin{subfigure}{0.33\textwidth}
    \includegraphics[width=\textwidth]{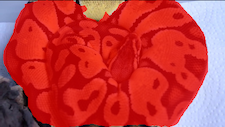}
\end{subfigure}%
\begin{subfigure}{0.33\textwidth}
    \includegraphics[width=\textwidth]{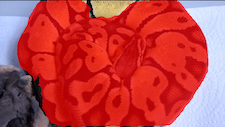}
\end{subfigure}
\caption{}
\end{subfigure}
%\vspace{-1em}
\captionsetup{labelfont=bf}
\caption{Qualitative results showing better separation of background \vs novel class for our multiscale memory decoding compared to our multiscale baseline. (a) Five-shot support set. (b) Multiscale Baseline. (c) Multiscale multigrid memory decoding (ours). The support groundtruth and query predictions are marked in red. Video results are in the appendix.}

\label{fig:qualimages}
\vspace{-1em}
\end{figure*}
Table~\ref{table:ablation_decoder_fsvos} shows the comparison for the FS-VOS task. It is seen that the multiscale query decoding, \textit{$+$MQuery-Stacked}, which outputs a compact representation from correlation tensors can degrade the results over the folds mean with respect to the \textit{Multiscale Baseline}. In contrast, we see that by working with a representation that preserves the spatiotemporal dimension, \textit{$+$MMemory-Stacked}, improvement occurs on the most challenging first fold by 1.3\%. 
%\textcolor{red}{You jump right into the 3rd vs 2nd line of the table without addressing the 1st vs 2nd: Bad.} 

Due to the differences between fully-supervised AVOS and FS-VOS, we face different challenges that do not occur in the AVOS task, \eg the potential existence of erroneous correlations at different scales, or confusing support examples. 
%\textcolor{red}{Challenges such as?} 
We overcome these challenges  with different variations on our multiscale memory decoding and ultimately outperform \textit{$+$MQuery-Stacked} by a considerable margin, as shown in the subsequent rows of Table~\ref{table:ablation_decoder_fsvos} and detailed below.

Additionally, we performed an ablation on the number of memory entries, $N$, used in our memory decoding. Table~\ref{table:nmem} compares $N=2$, a minimal value to capture both the novel class and background, \vs an over-complete set, $N=20$. The results generally support that use of an over-complete set to improve the separation of background and novel classes. Additional details are in the appendix. %\textcolor{red}{Would be good to a sweep of N, not just two extremes.}
%A detailed discussion and consideration of along with ablation on the fully supervised actor/action segmentation is left to the supplement.}
%The over-complete set captures different parts of the novel class and background and thereby improves the segmentation. In the supplement, we show visualisations of the attention maps that demonstrate how it is attending to different parts of the object and reveals the temporally consistent nature of them. We also ablate multiscale memory decoding in the fully supervised actor/action segmentation task provided in the supplement.}
%\textcolor{red}{This paragraph would be a good think to move entirely to the supplement.}

%with the visualisations of the attention maps, it also shows the temporal consistency of these maps across the video.

%\begin{table}[t!]
%\centering
%\small
%\begin{tabu}{@{}lccc@{}}
%\tabucline[1pt]{-}
%\multirow{2}{*}{Method} & \multicolumn{3}{c}{mIoU}\\ \cmidrule(r{2pt}){2-4}
%& DAVIS16~\cite{Perazzi2016} & MoCA~\cite{lamdouar2020betrayed} & YTBO~\cite{prest2012learning} \\ \tabucline[1pt]{-}
%MQuery-Stacked & 81.3 & 77.4 & 76.8 \\
%MMemory-Stacked & 83.5 & \textbf{80.3} & \textbf{78.2}\\
%MMemory-Multigrid & \textbf{83.7} &  79.3 & 78.1\\
%\tabucline[1pt]{-}
%\end{tabu}
%\captionof{table}{Ablation study on the AVOS task with three datasets reporting mean intersection over union using Video-Swin backbone. MQuery is multiscale query learning, MMemory is multiscale memory learning.}
%\label{table:ablation_decoder_avos}
%\captionsetup{labelformat=empty}
%\vspace{-1em}
%\end{table}

\begin{table}[t!]
\centering
\small
\begin{tabu}{@{}lccccc@{}}
\tabucline[1pt]{-}
\multirow{2}{*}{Method} & \multicolumn{5}{c}{mIoU}\\ \cmidrule(r{2pt}){2-6}
& 1 & 2 & 3 & 4 & Mean \\ \tabucline[1pt]{-}
$N=2$ & 49.6 & 70.3 & 62.9 & \textbf{65.1} & 61.98 \\
$N=20$ & \textbf{51.5} & \textbf{70.6} & \textbf{63.0} & 64.6 & \textbf{62.4}\\ \tabucline[1pt]{-}
\end{tabu}
\captionsetup{labelfont=bf}
\captionof{table}{Ablation study on the number of memory entries in our multiscale memory decoding on YouTube-VIS FS-VOS four folds with a five shot support set.}
\label{table:nmem}
\vspace{-1.2em}
\end{table}

\textbf{Information exchange across scales.}
%we compare four variants: (i) our baseline without a transformer decoder~\cite{min2021hypercorrelation}, (ii) the multiscale transformer with learnable queries that pools the spatiotemporal dimension following~\eqref{eq:target}, which we call \textit{Query}, (iii) our multiscale transformer that preserves the spatiotemporal dimensions,~\eqref{eq:st}, and follows a stacked information flow across scales as shown in Fig.~\ref{fig:multigrid} (left), which we call \textit{Stacked}
We now compare bidirectional, \eqref{eq:mma}, vs. stacked, \eqref{eq:sma}, information exchange across scales. Since information exchange across the spatiotemporal dimension is considered in both, we compare within multiscale memory decoding. Table~\ref{table:ablation_decoder_fsvos} shows that for FS-VOS the bidirectional information exchange ($+$\textit{MMemory-Multigrid}) provides consistent improvement over the stacked coarse-to-fine ($+$\textit{MMemory-Stacked}) across three of the four folds. Our hypothesis for the improvement is that it ameliorates erroneous correlations that occur in either the coarse or fine scales. To investigate, we show the relative robustness of the two schemes to noise perturbations averaged over all folds in Table~\ref{table:robustness}. We consider three types of noise applied to the normalized images, (i) Gaussian additive with 0.1 standard deviation, (ii) salt and pepper with 0.2\% of the original pixels values set to $0/1$ and (iii) speckle as multiplicative standard normal noise, $x \leftarrow  x + x*y_{\text{noise}}$, where $x$ is the original signal and $y_{\text{noise}}$ is standard normal noise. For all three types of noise, the bidirectional yields superior results, in support of our hypothesis, with up to 1\% gain.  %Across the three types of noise configurations, multigrid processing shows more robustness to noise than stacked processing with up to 1\% gain.

In contrast, Table~\ref{table:ablation_avos} shows the same ablation on AVOS, where it is found that the stacked is better on two of the three datasets. This result provides an interesting insight: When using input correlation tensors, as in FS-VOS, exchanging information in a bidirectional way is more beneficial compared to direct operation on the feature maps, as in AVOS. %This difference likely is due to the fact that direct feature comparisons do not require robustness to spurious correlations that our bidirectional exchange supplies.

\textbf{Meta-learning multiscale comparator transformer.}
We next study the benefits of meta-learning our multiscale comparator. 
%\textcolor{red}{Now, explain this experiment with respect to each relevant row in the table. BTW, to make this explanation coherent, you will want to refer to particular equations, which emphasizes the need to provide one for the adaptive k-shot. What currently follows remains difficult to follow.}
We compare our full MMC Transformer to conventional top-down multiscale processing using convolutional modules (\ie \textit{Multiscale Baseline}). 
%\textcolor{red}{If the results in the table literally are using \cite{min2021hypercorrelation}, then add the citation to the table. If not, then its confusing.} 
Table~\ref{table:ablation_decoder_fsvos} shows that our model (\textit{$+$MMemory-Multigrid}) improves over \textit{Multiscale Baseline} in the folds average and especially in the hardest fold (fold one). Since the core of our approach is the muliscale multigrid memory decoding, we show qualitative results on YouTube-VIS for (\textit{$+$MMemory-Multigrid}) in comparison to the \textit{Multiscale Baseline} in Fig.~\ref{fig:qualimages}. It is seen that our approach can capture objects of various sizes (\eg the snake as a big object), and can better separate the novel class from cluttered background (\eg person). 

Still, in the last fold \textit{$+$MMemory-Multigrid} degrades performance compared to \textit{Multiscale Baseline}; we hypothesize this results from confusing examples in the support set where our decoding might exacerbate the issue as it tries to better separate the novel class and the background. To overcome this problem we add an adaptive K-shot scheme, ~\eqref{eq:adaptive_kshot}, that reweighs the predictions from different shots to reduce the confusion from these examples (\textit{$+$MMemory-Multigrid$\dagger$}), which leads to improved results with respect to the baseline over all folds. Additionally, Table~\ref{table:robustness} shows the robustness of our method in comparison to the multiscale baseline even without the adaptive K-shot scheme. 
%\textcolor{red}{Previous sentence is overly redundant with what was presented when Table 6 was first introduced; best to remove here.} 
%\textcolor{red}{Assuming these qualitative results include the boundary refinement module, then its bad to introduce them here where the discussion has not yet considered that component. On the other hand, if the results in figure do not include boundary refinement, then you need to state that in the caption and explain why you are not showing results on your full approach.}
%Additionally, we show in Table~\ref{table:robustness} our results when exposing the model to different noise mechanisms applied on the input image to evaluate the robustness of our multiscale processing approach. It shows that our method outperforms the conventional multiscale baseline across all noise configurations using the mIoU averaged over the four folds and using five-shot support set, with up to 1.5\% gain.

%Table~\ref{table:ablation_a2d_fsvos} shows results on few-shot actor/action segmentation. We compare our conventional multiscale baseline~\cite{min2021hypercorrelation} and our MMC transformer on Common A2D. A greater improvement is seen with respect to the baseline than in object segmentation ablation, with up to 7\% gain in the five-shot scenario. Additionally, these results demonstrate the flexibility of our multiscale comparator, as it can operate beyond FSVOS to actor/action segmentation. 
%We provide Common A2D qualitative results in the supplement. 

\begin{table}[t!]
\centering
\small
\begin{tabu}{@{}lccc@{}}
\tabucline[1pt]{-}
Method & Gaussian & Salt \& Pepper & Speckle\\ \tabucline[1pt]{-}
Multiscale Baseline & 53.1 & 60.3 & 27.7 \\ \hline
$+$ \textbf{MMemory-Stacked} & 53.8 & 60.0 & 28.3 \\
$+$ \textbf{MMemory-Multigrid} & \textbf{54.2} & \textbf{61.0} & \textbf{29.2} \\ \tabucline[1pt]{-}
\end{tabu}
\captionsetup{labelfont=bf}
\captionof{table}{Robustness analysis of different scale exchange schemes when exposed to noise corruptions, reporting mIoU averaged over four folds for YouTube-VIS FS-VOS with five shot support.} %\textcolor{red}{Don't think we need the baseline here and its confusing.}}% \textcolor{red}{Noise configuration details are left to the supplement.}}
\label{table:robustness}
\vspace{-1.7em}
\end{table}

\textbf{Boundary refinement.}
Sole reliance on correlation tensors and ignoring the original query features can degrade boundary precision~\cite{hong2022cost}. To document how our boundary refinement module, \eqref{eq:bdry}, combats this challenge, we ablate the module, including its associated loss. %our proposed boundary refinement module and boundary loss that uses the refined features for the query video and is guided by the correlations output from our MMC Transformer. In 
Table~\ref{table:ablation_decoder_fsvos} shows that the module indeed improves segmentation quality (\textit{$+$MMemory-Multigrid$\ddagger$}). Additional ablation results for this module are provided in the supplement.%, where we compare the simple baseline with the additional boundary refiment \vs our MMCTransformer with boundary refinement.

\vspace{-0.5em}
\section{Conclusion}
\vspace{-0.5em}
\label{sec:conc}
We have presented MMC Transformer that exchanges information across scales on the full feature maps, rather than a compressed representation, and performs support-query comparisons while reducing the impact from erroneous correlations. 
%We meta-learn MMC along with a memory module to better separate a novel class from the background. 
It has been applied to the FS-VOS and AVOS tasks, where it outperforms the state of the art and baselines. %demonstrating the strength of our approach. %A limitation of our method for few-shot video object segmentation is its reliance on 2D backbone features, whereas in the video domain 3D spatiotemporal features may improve discriminability. We leave it for future work to explore spatiotemporal models as backbones and investigate FSVOS benchmarks that reflect the fact that certain semantic categories can exhibit different motion patterns (\eg four legged \vs two legged mammals \vs reptile motion).

%We leave it for future work to explore tasks that are beyond segmentation like common action detection~\cite{yang2021few}.
%approach that uses global and local temporal constraints to improve the accuracy of FS-VOS. These constraints are enforced as losses during learning, with the global addressing weight consistency across a video and the local enforcing region proportion consistency. Our approach outperforms state-of-the-art alternatives on a standard benchmark. We also introduced two novel benchmarks. The first, PASCAL-to-MiniVSPW, better captures realistic domain shift scenarios than extant FS-VOS benchmarks. The second, MiniVSPW-to-MiniVSPW, addresses the problem with non-exhaustive annotations that were provided in YouTube-VIS by providing annotations that label all occurrences of each novel semantic category. Our approach also shows state-of-the-art performance on these new benchmarks. %\textcolor{red}{Assuming space, we should say more here.}

\section{Appendix}
\subsection{Supplemental video}\label{sec:video}

We include an accompanying  video at \url{https://youtu.be/7LdnHjxD6JQ}. In this video, we show six examples for few-shot video object segmentation on YouTube-VIS~\cite{chen2021delving} and two examples for fully supervised automatic video object segmentation on MoCA~\cite{lamdouar2020betrayed}.
%and two examples for few-shot common action localisation on A2D~\cite{yang2021few}. 
Additionally, we show the attention maps output from the different memory entries for the finest scale across the entire input clip. The video is in MP4 format and is approximately seven minutes long. Layouts for each case are described in detail followed by the example video samples. The codec used for the realization of the provided video is H.264 (x264).

\subsection{Extension to actor/action segmentation}\label{sec:extratasks}
To confirm the versatility of our approach beyond video object segmentation we evaluate on the related task of actor/action segmentation. 

For fully supervised actor/action segmentation, similar to automatic video object segmentation, we use the feature pyramid from the backbone as direct input. Moreover, we modify the final convolutional layer in the segmentation decoder to output the 43 actor-action class pairs that appear in the A2D dataset. 

For few-shot actor/action segmentation, the support set becomes a trimmed video, unlike the few-shot video object segmentation with single image support examples, and the target query set has an untrimmed video. Thus, the one-way $K$-shot task has $K$ trimmed video-label pairs $\mathcal{S}=\{\mathsf{I}^{(s)}_k, {\mathsf{M}_k}^{(s)}\}_{k=1}^K$, where superscript $s$ stands for the support set. The binary segmentation mask, ${\mathsf{M}_k}^{(s)}$, is for one frame in the trimmed video for an actor/action class. The video-label pairs are of dimensions $\mathsf{I}_k \in \mathbb{R}^{T \times H \times W \times 3}$ and $\mathsf{M}_k \in \mathbb{R}^{H \times W}$. The few-shot actor/action segmentation model takes as input support-query correlations tensors between two backbone features. The first backbone captures spatial features from the reference frame in the clip, while the second backbone captures spatiotemporal features from the entire clip. The correlation tensors from both backbones are concatenated on the channel dimension. We find this change necessary because, unlike semantic segmentation, action segmentation depends not only on spatial features, but also spatiotemporal features (\eg motion). %During inference, We use a temporal sliding window over the untrimmed target query video and generate clips that are used to predict the segmentation. 

\subsection{Experiment design details}
\label{sec:impdetails}

In this section, we detail the experiment setup and implementation details that were not described in the main submission due to space considerations.

\subsubsection{Evaluation protocols and datasets}
We start with describing the datasets we use in both the \textit{few-shot} and \textit{fully supervised} tasks.

\textbf{YouTube-VIS FS-VOS~\cite{chen2021delving} (\textit{few-shot})} In the few-shot video object segmentation task we use the YouTube-VIS~\cite{chen2021delving} dataset that is split into four folds, each with 30 training classes and 10 test classes. The dataset contains high resolution videos with resolution of 720p. Evaluation is performed as the average over five runs, as elsewhere~\cite{chen2021delving}; both mean intersection over union, $\mathcal{J}$, and mean boundary accuracy, $\mathcal{F}$, are reported. %\textcolor{red}{What is the resolution? If its variable, then can we provide upper and lower bounds?}

\textbf{Common A2D~\cite{yang2021few} (\textit{few-shot})} In few-shot common action localisation we use Common A2D that has 43 classes split into 33, five and five classes for the training, validation and test splits, resp. The dataset has video frames with resolution 320p. A difference in our use of the dataset with respect to previous work~\cite{yang2021few}, is that we use an additional segmentation mask for one frame in the trimmed support video, whereas previous approaches did not require a segmentation mask. In our case, since we use correlation tensors we need to define the foreground action; otherwise, even the static background from both support and target query can have high correlations. We leave it for future work to investigate weaker forms of supervision. To evaluate on this dataset we use mean intersection over union, similar to state-of-the-art approaches on this benchmark~\cite{yang2021few}. %\textcolor{red}{What is the resolution? If its variable, then can we provide upper and lower bounds?}

\textbf{DAVIS'16~\cite{Perazzi2016} (\textit{fully supervised})} consists of 50 video sequences with a total 3455 annotated frames of spatial resolution 480p and 720p, with high quality dense pixel-level annotations (30 videos for training and 20 for testing). For quantitative evaluation on this dataset, we follow the standard evaluation protocol provided with the dataset. We report results as mean Intersection over Union (mIoU), $\mathcal{J}$, and mean boundary accuracy, $\mathcal{F}$, using 480p resolution frames. We perform postprocessing similar to state-of-the-art methods on this dataset; specifically, we use multiscale inference~\cite{zhen2020learning}. In the multiscale inference we use inputs at different scales then average predictions. The scale factors we use are $\{0.7, 0.8, 0.9, 1.0, 1.1, 1.2\}$. Note that we use postprocessing only for this dataset, as is the norm; none of our other reported results entail postprocessing. %\textcolor{red}{What is the resolution?}

\textbf{YouTube Objects~\cite{prest2012learning} (\textit{fully supervised})} contains 126 videos with more than 20,000 frames. All videos are used for testing; the dataset does not include training and validation subsets. Following its protocol, we use mean intersection over union (mIoU) to measure performance per category and the mean over all categories. The dataset has video frames with resolution 720p.%\textcolor{red}{What is the resolution? If its variable, then can we provide upper and lower bounds?}

\textbf{MoCA~\cite{lamdouar2020betrayed} (\textit{fully supervised})} consists of videos of moving camouflaged animals with more than 140 clips of resolution 720p across a diverse range of animals. All videos are used for testing; the dataset does not include training and validation subsets. This is the most challenging motion segmentation dataset currently available, as in the absence of motion the camouflaged animals are almost indistinguishable from the background by appearance alone (\ie colour and texture). The groundtruth is provided as bounding boxes, and we follow previous work~\cite{yang2021self} in removing videos that contain no predominant target locomotion, to evaluate on a subset of 88 videos. We evaluate using mIoU with the largest bounding box on the predicted segmentation and also report success rate with varying IoU thresholds, $\tau \in \{0.5,...,0.9\}$.

\textbf{A2D~\cite{xu2015can} (\textit{fully supervised})} consists of 3782 videos from YouTube that were annotated for nine different actors and seven actions, the final valid categories of actor-action pairs constitute 43 pairs. Following its protocol we evaluate mean intersection over union on these actor-action categories. The dataset has video frames with resolution 320p and it contains training and testing subsets, where we evaluate on the test set.
%\textcolor{red}{Are there training and validation subsets, if not, then say so. What is the resolution? If its variable, then can we provide upper and lower bounds?}

\subsection{Implementation details}
\textbf{Few-Shot video dense prediction.} For few-shot video object segmentation, we use the ResNet-50~\cite{HeZRS16} backbone, similar to a state-of-the-art FS-VOS method~\cite{chen2021delving}. For  actor/action segmentation, we use both ResNet-50 to extract spatial features and X3D~\cite{feichtenhofer2020x3d} to extract spatiotemporal features. In both tasks we do not train the weights of the backbone and the features that are used to compute the correlation tensors are extracted as follows. We extract features at the end of each bottleneck before the ReLU activation in the last three stages and concatenate the correlation tensors (with same spatial dimensions) along the channel dimension; this operation results in a three scale hypercorrelation pyramid, $L=3$. 

Our MMC Transformer is constructed as seven decoding layers with separate weights, each corresponding to one bidirectional communication exchange across scale. It is composed of multihead attention with eight heads and operates on features with 128 channels. In the bidirectional scale exchange we go through coarse-mid-fine-mid-coarse-mid-fine in that order, which results in the final enhanced feature maps at the finest scale. The boundary refinement module is constructed as 12 transformer layers and splits the input query video per frame into non-overlapping patches of size $8\times8$. The boundary refinement is pretrained with self supervision~\cite{caron2021emerging} to help extract better boundaries.
%; these weights are fixed during the meta-training, only to learn the boundary and segmentation prediction heads. 
Recall that our boundary refinement module entails multiple attention blocks followed by two parallel convolutional prediction heads to produce the segmentation mask and its boundary. The attention blocks are pretrained with self supervision~\cite{caron2021emerging} and consequently are frozen. Subsequently, the final segmentation and boundary prediction heads are meta-trained and guided by the features from our multiscale transformer decoding.  %\textcolor{red}{From semi-colon to end of sentence is unclear. Are you saying that following boundary pretraining the boundary module weights are frozen and subsequently metatraining proceeds to set all other weights, including those of the prediction head? } These heads use the enhanced correlation features output from our decoder as a guide to segment the extracted query features. %We train our models on a single NVIDIA RTX A6000 GPU. 

\textbf{Fully supervised video dense prediction.} For the automatic video object segmentation and fully supervised actor/action segmentation we use the same detailed architecture. For the backbones in the main submission, we use Video-Swin~\cite{liu2022video}. Here, in the supplement, we also report results with the ResNet-101 backbone for the AVOS task, similar to the state-of-the-art AVOS approaches~\cite{zhou2020motion,ren2021reciprocal}. We extract features from the last four stages to construct a feature pyramid with four scales. For our multiscale memory transformer decoder we use eight layers with multihead attention of eight heads and features channel as 384. Our baseline uses the same settings, but with nine decoder layers reflecting the last three scales with a stacked decoding scheme~\cite{cheng2021masked}. Additionally, the transformer decoder, similar to best practices reported in earlier semi-supervised VOS~\cite{cheng2021mask2former,wang2021end}, is comprised of consecutive self and cross attention modules, and is used in both our \textit{Multiscale Baseline (MQuery Stacked)} as well as our multiscale memory transformer decoder. 

\subsection{Inference and training details.}
\textbf{Few-Shot video dense prediction.} For the few-shot video object and action segmentation we train our model with 50 epochs on YouTube-VIS, while we use 70 epochs for A2D. We use the same hyperparameters for both our approach and the baseline, where we use AdamW~\cite{loshchilov2018decoupled} with a learning rate of $1\times10^{-3}$ and weight decay of $1 \times 10^{-4}$. We use data augmentation instantiated as random horizontal flipping as well as random resize and rotate within (-15, 15) degrees, as elsewhere~\cite{chen2021delving}. We use batch size 19 for YouTubeVIS FS-VOS and batch size seven for Common A2D. We follow standard few-shot segmentation and few-shot video segmentation practice of assigning the novel class objects that exist in training images to background~\cite{shaban2017one}. During training on YouTube-VIS FS-VOS, we randomly sample two frames every fifth frame from the query set video and randomly sample one image from another video for the support set. In Common A2D, we use 25 frame trimmed clips subsampled uniformly into five frames for the support and query sets during training. During inference, in YouTube-VIS FS-VOS we infer over the entire video. While in Common A2D during inference, we use 25 frame clips in a temporal sliding window over the entire untrimmed query video. We train our models on a single NVIDIA RTX A6000 GPU. %\textcolor{blue}{add data augmentations}

\textbf{Fully supervised video dense prediction.} We train our models for the fully supervised tasks using $1 \times 10^{-4}$ as learning rate for the transformer decoder and pixel decoding, while we use a learning rate of $1 \times 10^{-6}$ for the backbone and weight decay of $1 \times 10^{-4}$. We use a polynomial learning rate decay with power 0.9 and data augmentation instantiated as random horizontal flipping as well as random resize, as elsewhere~\cite{ren2021reciprocal}. In both video object and action segmentation we train the models using a combination of the focal loss~\cite{lin2017focal} and dice loss~\cite{milletari2016v} that are weighted equally. In the AVOS task, we sample six consecutive frames for the training; for actor/action segmentation we input video of 25 frames subsampled into six frames with the center frame as the one with a groundtruth mask. In both tasks we use $N=6$ for our multiscale memory entries. Our AVOS models are trained on both YouTube-VOS and DAVIS'16 training-sets, following state-of-the-art approaches~\cite{zhou2020motion} for 210,000 iterations. While in the actor/action segmentation we train on the A2D training set for approximately 46,000 iterations. Both our approach and the baseline are pretrained using weights trained on MS-COCO following~\cite{carion2020end}. Training employs a single NVIDIA RTX A6000 GPU.
%\textcolor{blue}{add data augmentations + learning rate decay}

\subsection{Additional empirical results}
\label{sec:results}

In this section, we provide additional empirical results in comparison to the state of the art in AVOS and actor/action segmentation. Moreover, we provide experiments to understand better the multiscale memory learning and confirm the motivation of our proposed multigrid formulation.
%\textcolor{red}{Force all figures to appear in the order they are introduced in the text.}

\subsubsection{Comparison to state of the art}
\textbf{AVOS additional metrics.} Here, we provide the full evaluation results of our AVOS model with multiscale memory transformer decoder across the three datasets considered in the main submission, MoCA, YouTube Objects and DAVIS'16. For the sake of fair comparison to the state-of-the-art AVOS methods~\cite{zhou2020motion,ren2021reciprocal} we report using the ResNet-101 backbone. 

Our results on the most challenge camouflaged moving animals dataset in Table~\ref{tab:moca:sota} shows that our approach, without the use of extra optical flow input, outperforms the state of the art with a considerable margin of up to 2.4\% in mIoU and 4.6\% in mean success rate. Moreover, in Table~\ref{tab:ytbobj:sota} we show the results on yet another large-scale AVOS dataset, YouTube Objects, which confirms the superiority of our approach even with the weaker backbone than that considered in the main submission, \ie ResNet-101. 

In Table~\ref{tab:davis16:sota} we report our results on DAVIS'16, with and without postprocessing, where we use a simple post processing strategy of multiscale inference, as detailed above. It is seen that our approach, without optical flow and with postprocessing, outperforms the state-of-the-art AVOS approaches that use only RGB images as input. For approaches that use extra optical flow input, when we compare to the previous second best approach, MATNet~\cite{zhou2020motion} we report on par mIoU and better boundary accuracy without the extra optical flow input. However, the best approach that uses both optical flow and RGB images, RTNet~\cite{ren2021reciprocal}, performs better than our model on DAVIS'16; although, we outperform it on MoCA with a considerable margin of 5.9\%. The relative success of RTNet on DAVIS'16 can be attributed to its use of the extra pretraining saliency segmentation dataset, DUTS~\cite{wang2017learning}, which we do not use. These results align with previous conclusions that DAVIS'16 is a static biased dataset and focuses more on information conveyed from single images~\cite{kowal2022deeper} than motion information. Yet, with the improved backbone, Video-Swin, our model outperforms RTNet without optical flow input or this extra dataset.

%Js: (0.803). SR at 0.5: (0.911), 0.6: (0.882), 0.7: (0.826), 0.8: (0.705), 0.9: (0.448),
%Js: (0.666). SR at 0.5: (0.739), 0.6: (0.704), 0.7: (0.644), 0.8: (0.543), 0.9: (0.321),
\begin{table*}[t]
  \centering
  \aboverulesep=0ex
   \belowrulesep=0ex
   \begin{adjustbox}{max width=0.8\textwidth}
    \begin{tabular}{@{}cc|ccc|ccc@{}}
    \toprule
    \multicolumn{2}{c|}{\multirow{2}{*}{Measures}} & \multicolumn{3}{c|}{Uses RGB+Flow} & \multicolumn{3}{c}{Uses RGB only} \\
    % \cmidrule(r){3-7}
    \multicolumn{2}{c|}{}                   & COD (two-stream)~\cite{lamdouar2020betrayed}  &   MATNet~\cite{zhou2020motion}  &   RTNet~\cite{ren2021reciprocal}  &  COSNet~\cite{lu2019see} & \textbf{Ours} & \textbf{Ours$\dagger$}  
    \\ 
    \midrule
    % --------------------------------------------------------------------------------------------
    \multirow{1}{*}{$\mathcal{J}$ Mean $\uparrow$}          &          & 55.3 &  64.2   &  60.7   &  50.7 & 66.6 & \textbf{80.3}\\
    % --------------------------------------------------------------------------------------------
    \cmidrule{1-8}
    % --------------------------------------------------------------------------------------------
    \multirow{6}{*}{Success Rate $\uparrow $}       &   $\tau = 0.5$     &  0.602 &  0.712   & 0.679    &  0.588  & 0.739 & \textbf{0.911}\\
                                                    &   $\tau = 0.6$     &  0.523 &  0.670   & 0.624    &  0.534  & 0.704 & \textbf{0.882}\\
                                                    &   $\tau = 0.7$     &  0.413 &  0.599   & 0.536    &  0.457  & 0.644 & \textbf{0.826}\\
                                                    &   $\tau = 0.8$     &  0.267 &  0.492   & 0.434    &  0.337  & 0.543 & \textbf{0.705}\\
                                                    &   $\tau = 0.9$     &  0.088 &  0.246   & 0.239    &  0.167  & 0.321 & \textbf{0.448}\\
                                                    &   $SR_{mean}$      &  0.379 &  0.544   & 0.502    &  0.417  & 0.590 & \textbf{0.754}\\
    % --------------------------------------------------------------------------------------------
    %\cmidrule{1-12}
    % --------------------------------------------------------------------------------------------
    %       $\mathcal{T}$                                &    Mean      & -/18.5  &  /21.6   &   -           &  -          & -/44.5 &  -/31.9   & -/33.7 &   -       & - & /82.7\\
    \bottomrule
    \end{tabular}
    \end{adjustbox}
  \caption{Results of moving camouflaged object segmentation on MoCA dataset with best overall results in \textbf{bold}. Results shown as mean Intersection over Union (mIoU) and localization success rate for various tresholds, $\tau$. We show our results with the standard ResNet-101 backbone and Video-Swin backbone, indicated by $\dagger$.
  %as standard~\cite{yang2021self}. 
  }
  \label{tab:moca:sota}
\end{table*}

%0.861 | 0.967  | 0.050 | 0.872 | 0.951  | 0.027
%0.838 | 0.938  | 0.057 | 0.865 | 0.944  | 0.029
%0.819  |  0.917   |  0.041  | 0.841  |  0.922   |  0.035
%0.823  |  0.919   |  0.029  | 0.839  |  0.919   |  0.031
\begin{table*}[t]
  \centering
  \aboverulesep=0ex
   \belowrulesep=0ex
   \begin{adjustbox}{max width=\textwidth}
    \begin{tabular}{@{}cc|cccc|ccccccc@{}}
    \toprule
    \multicolumn{2}{c|}{\multirow{2}{*}{Measures}} & \multicolumn{4}{c|}{Uses RGB+Flow} & \multicolumn{6}{c}{Uses RGB only} \\
    % \cmidrule(r){3-12}
    \multicolumn{2}{c|}{}  & EPO+~\cite{akhter2020epo}  &   MATNet~\cite{zhou2020motion}  &   RTNet~\cite{ren2021reciprocal}  &  FSNet ~\cite{ji2021full}   & AGS~\cite{wang2019learning}   & COSNet~\cite{lu2019see} &  AGNN~\cite{wang2019zero}  & ADNet~\cite{yang2019anchor} & DFNet~\cite{zhen2020learning}   & \textbf{Ours} & \textbf{Ours$\dagger$}  \\
    \midrule
    % --------------------------------------------------------------------------------------------
    \multirow{3}{*}{$\mathcal{J}$}          &    Mean $\uparrow $      & -/80.6 &  -/82.4   &  84.3/85.6   &  82.1/83.4 & -/79.7 &  -/80.5 & 78.9/80.7 & 78.26/81.7 &  -/83.4 & 81.9/82.3& 83.8/\textbf{86.1} \\
                                            &    Recall  $\uparrow $  & -/95.2 &  -/94.5   &   -/96.1     &  -         & -/91.1 &   -/94.0 & -/94.0 &   -        &  -  & 91.7/91.9 & 93.8/\textbf{96.7} \\
                                            &    Decay $\downarrow $    & -/0.02 &  -/5.5    &   -          &  -         & -/\textbf{0.0}  &   -/\textbf{0.0}  & -/0.03 &   -        &  - & 0.04/0.03 & 0.05/0.05\\
    % --------------------------------------------------------------------------------------------
    \cmidrule{1-13}
    % --------------------------------------------------------------------------------------------
    \multirow{3}{*}{$\mathcal{F}$}          &   Mean    $\uparrow $   & -/75.5 &  -/80.7   & 83.0/84.7    &  83.0/83.1  & -/77.4 &  -/79.4  & -/79.1 & 77.1/80.5  & -/81.8  & 84.1/83.9 & 86.5/\textbf{87.2}\\
                                            &   Recall   $\uparrow $  & -/87.9 &  -/90.2   &-/93.8        &  -          & -/85.8 &  -/90.4  & -/90.5 &  -         &  - & 92.2/92.0 & 94.4/\textbf{95.1}\\
                                            &   Decay  $\downarrow $    & -/0.02 &  -/4.5    &  -           &  -          & -/\textbf{0.0}  &  -/\textbf{0.0}   & -/0.03 &  -         &  - & 0.04/0.03 & 0.03/0.03\\
    % --------------------------------------------------------------------------------------------
    %\cmidrule{1-12}
    % --------------------------------------------------------------------------------------------
    %       $\mathcal{T}$                                &    Mean      & -/18.5  &  /21.6   &   -           &  -          & -/44.5 &  -/31.9   & -/33.7 &   -       & - & /82.7\\
    \bottomrule
    \end{tabular}
    \end{adjustbox}
  \caption{Results on DAVIS'16 validation set. For those using post processing (\eg conditional random fields~\cite{akhter2020epo, zhou2020motion, ren2021reciprocal, ji2021full, wang2019learning, lu2019see, wang2019zero,zhen2020learning}, instance pruning~\cite{yang2019anchor, zhen2020learning}, multiscale inference~\cite{zhen2020learning}), results shown as $x/y$, with $x$ and $y$ results without and with post processing, resp. Boundary accuracy, $\mathcal{F}$, and mean Intersection over Union (mIoU), $\mathcal{J}$ are shown. %Best result of approaches using RGB and flow in red; best result of approaches using RGB only in blue. 
  We show our results with the standard ResNet-101 backbone and Video-Swin backbone, indicated by $\dagger$. Best results highlighted in \textbf{bold}.
  }
  \label{tab:davis16:sota}
\end{table*}

\begin{table*}[t!]
\centering
  \aboverulesep=0ex
   \belowrulesep=0ex
\begin{adjustbox}{max width=0.9\textwidth}
	    \begin{tabular}{@{}c|cccc|cccccc@{}}
		\toprule
		\multirow{2}{*}{Category} & \multicolumn{4}{c|}{Uses RGB+Flow} & \multicolumn{6}{c}{Uses RGB only} \\
		% \cmidrule(r){2-10}
		 & FSEG~\cite{jain2017fusionseg} &   LVO~\cite{tokmakov2017learning} & MATNet~\cite{zhou2020motion}  &   RTNet~\cite{ren2021reciprocal}  &  PDB~\cite{song2018pyramid}   & AGS~\cite{wang2019learning}   &  COSNet~\cite{lu2019see} & AGNN~\cite{wang2019zero}  &   \textbf{Ours}  & \textbf{Ours$\dagger$}\\ 
		\midrule
		Airplane(6)  & 81.7	& 86.2  		& 72.9  & 84.1          & 78.0  	& 87.7 & 81.1  & 81.1          & 88.2 & \textbf{88.2}\\
		Bird(6)      & 63.8	& \textbf{81.0} & 77.5  & 80.2          & 80.0  	& 76.7 & 75.7  & 75.9          & 75.6 & 79.8\\
		Boat(15)     & 72.3	& 68.5  		& 66.9  & 70.1          & 58.9		& 72.2 & 71.3  & 70.7          & 69.3 & \textbf{77.5}\\
		Car(7)       & 74.9	& 69.3  		& 79.0  & 79.5 & 76.5		& 78.6 & 77.6  & 78.1          & 78.3 & \textbf{86.8}\\
		Cat(16)      & 68.4	& 58.8  		& 73.7  & 71.8          & 63.0		& 69.2 & 66.5  & 67.9          & 80.2 & \textbf{83.5}\\
		Cow(20)      & 68.0	& 68.5  		& 67.4  & 70.1          & 64.1		& 64.6 & 69.8  & 69.7          & 73.6 & \textbf{77.2}\\
		Dog(27)      & 69.4 & 61.7  		& 75.9  & 71.3          & 70.1		& 73.3 & 76.8  & 77.4 & 76.4 &  \textbf{79.3} \\
		Horse(14)    & 60.4 & 53.9  		& 63.2  & 65.1          & 67.6		& 64.4 & 67.4  & 67.3          & \textbf{69.1} & 68.9 \\
		Motorbike(10)& 62.7 & 60.8  		& 62.6  & 64.6          & 58.3		& 62.1 & 67.7  & \textbf{68.3} &  66.0 &  68.1  \\
		Train(5)     & 62.2 & 66.3  		& 51.0  & 53.3          & 35.2		& 48.2 & 46.8  & 47.8          &  66.5  & \textbf{73.0} \\
		\midrule
		Mean     	 & 68.4 & 67.5 			& 69.0  & 71.0          & 65.4		& 69.7 & 70.5	 & 70.8			 &  74.3 & \textbf{78.2} \\
		\bottomrule
	\end{tabular}
\end{adjustbox}
\caption{Results of various object class segmentations on YouTube Objects. Results shown as mean Intersection over Union (mIoU) per category as well as average across all categories. Best results highlighted in \textbf{bold}. We show our results with the standard ResNet-101 backbone and Video-Swin backbone, indicated by $\dagger$.}
\label{tab:ytbobj:sota}
\end{table*}

\textbf{FS-VOS runtime analysis.}
We compare our multiscale memory comparator transformer in few-shot video object segmentation to DANet~\cite{chen2021delving} that proposed an online learning scheme that finetunes the backbone along with a many-to-many attention comparator. Their method reported 20 seconds per video on a 2080Ti GPU, although we did not have access to the same GPU we report results on a lower tier TITAN-X GPU which resulted in a runtime of 2.9 seconds per video on Youtube-VIS on average, which is approximately a 10x speedup. This fact confirms that with our multiscale comparator, there is no need to perform backbone finetuning during inference, which makes our approach computationally efficient and better in performance.

\subsubsection{Actor/action segmentation} In this section we demonstrate our results for actor/action segmentation in the few-shot and fully supervised settings. We provide these results  to show the potential to deploy our approach in different video understanding tasks, including not only object, but also action segmentation.

\textbf{Few-shot actor/action segmentation.} We build a simple multiscale baseline, similar to the few-shot image segmentation work~\cite{min2021hypercorrelation}, but meta-trained for the few-shot common action localisation sampled episodes. Initially, we used only X3D~\cite{feichtenhofer2020x3d} as our backbone \textit{(Multiscale Baseline)}, but we found it insufficient to capture necessary details for the actors. Thus, we resorted to a better multiscale baseline that uses both spatial (from ResNet-50) and spatiotemporal (from X3D) features and combines them \textit{(Multiscale Baseline)$\dagger$}.
%\textcolor{red}{DEFINE STATIC vs. DYNAMIC COMPARATORS REFERRED TO IN TABLE CAPTION.} 
In Table~\ref{table:ablation_a2d_fsvos}, we compare our best multiscale baseline with conventional top-down processing to our proposed MMC Transformer on the common A2D benchmark using mean intersection over union for both 1-shot and 5-shot tasks. The comparison shows that our MMC Transformer outperforms the baseline with a considerable margin $\approx$ 6\%. 

\begin{table}[t!]
\centering
\small
\begin{tabu}{@{}lcc@{}}
\tabucline[1pt]{-}
\multirow{2}{*}{Method} & \multicolumn{2}{c}{mIoU} \\ \cmidrule(r{2pt}){2-3}
& 1-shot & 5-shot \\ \tabucline[1pt]{-}
%Co-attention~\cite{siam2020weakly} & 43.3 & 44.8\\
%Single-scale Transformer~\cite{yang2021few} &  50.6 & 52.5\\ %\tabucline[1pt]{-}
Multiscale Baseline & 18.1 & 21.9\\ 
Multiscale Baseline$\dagger$ & 45.2 & 47.7\\ \hline
\textbf{MMCT (Ours)$\dagger$} &\textbf{51.9} & \textbf{54.5}\\ \tabucline[1pt]{-}
\end{tabu}
\caption{Comparisons of mIoU for few-shot actor/action segmentation with respect to our baseline on Common A2D and our baselines, with one-shot and five-shot support sets. $\dagger$ indicates the use of both spatial (ResNet-50) and spatiotemporal (X3D) backbones and fusing their correlation tensors.}
\vspace{-1.22em}
\label{table:ablation_a2d_fsvos}
\end{table}

\textbf{Fully supervised actor/action segmentation.} For the fully supervised benchmark we use a multiscale baseline similar to AVOS, \textit{(Multiscale Baseline (MQuery-Stacked))}, which was detailed in the main submission. In Table~\ref{table:ablation_a2d} we see again that our approach with multiscale memory decoding improves over the conventional multiscale baseline.

\begin{table}[t!]
\centering
\small
\begin{tabu}{@{}lc@{}}
\tabucline[1pt]{-}
Method & mIoU\\ \tabucline[1pt]{-}
%Co-attention~\cite{siam2020weakly} & 43.3 & 44.8\\
%Single-scale Transformer~\cite{yang2021few} &  50.6 & 52.5\\ %
\makecell[lt]{Multiscale Baseline\\ (MQuery-Stacked)} & 51.1\\ \tabucline[1pt]{-}
%Multiscale Baseline$\dagger$ & 45.2 & 47.7\\ \hline
\textbf{MMCT (Ours)} & \textbf{53.8}\\ \tabucline[1pt]{-}
\end{tabu}
\caption{Comparisons of mIoU for fully supervised actor/action segmentation with respect to our multiscale baseline on A2D.}
\vspace{-1.22em}
\label{table:ablation_a2d}
\end{table}

\subsubsection{Additional ablation studies}
\textbf{Boundary refinement ablation.} In the main submission we confirmed the benefit of the boundary refinement module when added to our MMC transformer model. In this section we do a further ablation of comparing our conventional multiscale baseline with the additional boundary refinement module without the use of our proposed multiscale memory transformer decoding.  Table~\ref{table:ablation_bdry_refine} shows that while the addition of boundary refinement improves the multiscale baseline, it still lags behind our full approach that further includes multiscale memory transformer decoding.

\begin{table}[t!]
\centering
\small
\begin{tabu}{@{}lccccc@{}}
\tabucline[1pt]{-}
\multirow{2}{*}{Method} & \multicolumn{5}{c}{mIoU}\\ \cmidrule(r{2pt}){2-6}
& 1 & 2 & 3 & 4 & Mean \\ \tabucline[1pt]{-}
Multiscale Baseline & 49.5 & 69.5 & 63.8 & 65.2 & 62.0 \\
\hspace{0.8em} Multiscale Baseline$\ddagger$ & 48	& 71.6	& 64.1	& 67.5	& 62.8\\ \hline
$+$ \textbf{MMemory-Multigrid$\ddagger$} & \textbf{52.2} & \textbf{73.8} & \textbf{65.7} & \textbf{68.7} & \textbf{65.1} \\ \tabucline[1pt]{-}
\end{tabu}
\captionof{table}{Ablation study on YouTube-VIS FS-VOS folds 1, 2, 3 and 4 using our MMC Transformer with a five shot support set. $\ddagger$ indicates an additional boundary refinement. MMemory indicates our multiscale memory transformer decoding scheme.
%5\textcolor{red}{Not a clean study: All you want to document is the benefit of multiscale memory decoding, the rest is noise or else would require much elaboration of each and every variant in the text. See also my comments in the text Re. this study.} 
}
\label{table:ablation_bdry_refine}
\captionsetup{labelformat=empty}
\vspace{-1em}
\end{table}

\textbf{GPU memory considerations.} Consideration of the number of parameters for our multiscale memory transformer decoding \vs the baseline without the augmented memory module, shows just a small increment. In particular, the parameters for our full approach \vs the baseline 28.6M and 28.1M, resp. Nonetheless, one of the limitations that can occur with multiscale memory transformer decoding relates to higher consumption of GPU memory. For the FS-VOS task this is less problematic since we operate on processed correlation tensors with lower dimensionality than what normally occurs in direct feature maps. Moreover, our transformer decoding in the few-shot uses cross-attention, which again has lower memory consumption than what self attention normally would require. However, when working on fully supervised automatic video object segmentation, higher memory consumption could be problematic. The multiscale memory transformer decoding design we follow is similar to our multiscale query learning baseline, \textit{Multiscale Baseline (MQuery Stacked)}. This decoding entails the use of consecutive cross and self attention blocks where the self attention is conducted on the queries. In the case of multiscale memory decoding, the queries are the detailed feature maps and performing self attention especially in the finest scale can incur significant memory consumption. 

We conduct experiments that show an efficient version of our multiscale memory decoding, where we drop a percentage of the tokens processed during self attention only in the finest scale. Then we use these tokens without self attention combined with the output from self attention as input to our cross attention. This operation is only performed in the finest scale and on a small percentage of the tokens; therefore, it has limited impact on the cross attention that is the heart of our decoding scheme. This modification is inspired by classical adaptive multigrid operations~\cite{briggs2000multigrid}, that only focuses on certain regions of interest in the finest scales computations for the sake of efficiency.  Figure~\ref{fig:mem_analysis} shows that without this modification increase of the input resolution leads to an exponential increase in the memory consumption with our multiscale memory decoding. Yet, when we use our efficient implementation it shows a considerable decrease in our memory consumption while maintaining on-par mean intersection over union on the challenging MoCA benchmark. We leave it for future work to investigate better adaptive mechanisms that select the region of interest to be processed at the finest scales in a guided fashion and to investigate its effect on other datasets.

\begin{figure*}[t!]
    \centering
    \resizebox{0.5\textwidth}{!}{
\begin{tikzpicture}
\begin{axis} [
    width=\axisdefaultwidth,
    height=6cm,
    ymin=0, ymax=50,
    xmin=100, xmax=450,
    ytick = {0, 10, 20, 30, 40, 50},
    xtick = {150, 200, 250, 300, 350, 400, 450},
    %title = \textbf{Memory Analysis},
    ylabel = GPU Memory in GiB,
    xlabel = Input Resolution,
    enlarge x limits = {value = .1},
    enlarge y limits={abs=0.8}, 
] 
            
\addplot[magenta,mark=o,mark options={line width=0.9pt,scale=1.2,solid},style=solid] coordinates {(150, 5.6) (200, 7.1) (250, 10.7) (300, 16.6) (350, 25.8) (400,39.7) (440, 48.0)};
\addplot[cyan,mark=x,mark options={line width=0.9pt,scale=1.2,solid},style=solid] coordinates {(150, 5.1) (200, 5.6) (250, 6.9) (300, 8.9) (350, 11.8) (400, 16.1) (440, 20.9)};

%\legend{MMemory, MMemory Adaptive}
\end{axis}
\end{tikzpicture}
}%
\resizebox{0.5\textwidth}{!}{
\begin{tikzpicture}
\begin{axis} [
    width=\axisdefaultwidth,
    height=6cm,
     ymin=50, ymax=100,
    xmin=100, xmax=450,
    ytick = {50, 60, 70, 80, 90, 100},
    xtick = {150, 200, 250, 300, 350, 400, 450},
    %title = \textbf{Memory Analysis},
    ylabel = mIoU,
    xlabel = Input Resolution,
    enlarge x limits = {value = .1},
    enlarge y limits={abs=0.8},
]       

\addplot[magenta,mark=o,mark options={line width=0.9pt,scale=1.2,solid},style=solid] coordinates {(150, 61.1) (200, 73.2) (250, 78.0) (300, 79.1) (350, 80.3) (400, 80.3) (440, 80.3)};
\addplot[cyan,mark=x,mark options={line width=0.9pt,scale=1.2,solid},style=solid] coordinates {(150, 61.6) (200, 73.5) (250, 78.2) (300, 79.2) (350, 80.2) (400, 80.2) (440, 80.5)};

\legend{MMemory, MMemory-Eff}

\end{axis}
\end{tikzpicture}
}
\caption{Analysis of memory consumption during inference and mean intersection over union for both Muliscale Memory learning (\textit{MMemory}) and an efficient version with lower memory consumption (\textit{MMemory-Eff}). The analysis is conducted on the automatic VOS task on the challenging MoCA benchmark.}
\label{fig:mem_analysis}
\end{figure*}

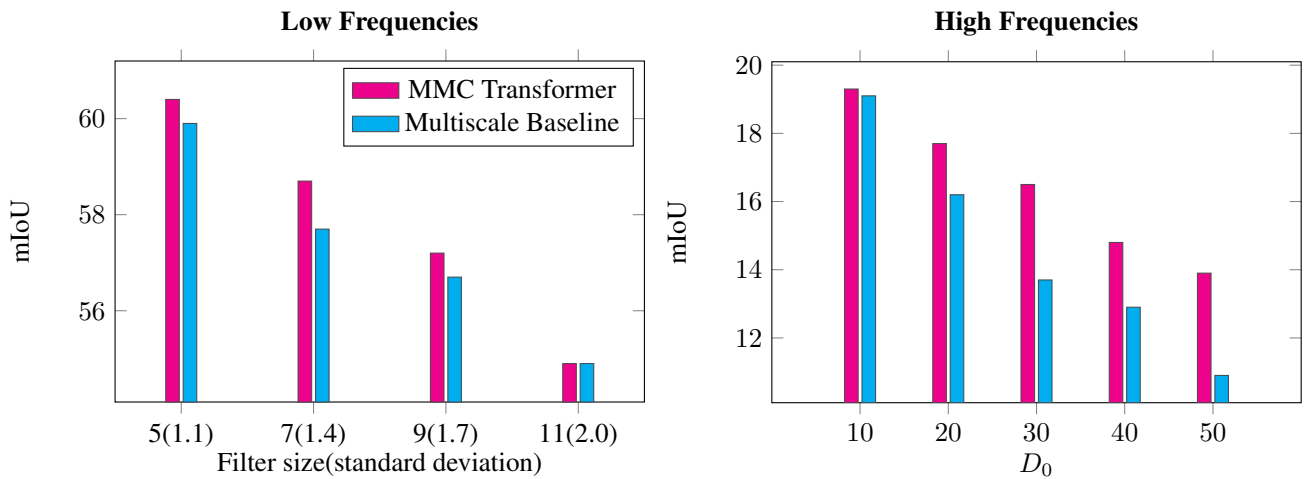
\begin{figure*}
    \centering
    \resizebox{0.5\textwidth}{!}{
\begin{tikzpicture}
\begin{axis} [
    width=\axisdefaultwidth,
    height=6cm,
    bar width = 5pt,
    ybar = .05cm,
    xmin = 4,
    xmax = 12,
    xtick={5, 7, 9, 11},
    xticklabels={{5(1.1)}, {7(1.4)}, {9(1.7)}, {11(2.0)}},
    title = \textbf{Low Frequencies},
    ylabel = mIoU,
    xlabel = Filter size(standard deviation),
    area legend,
    enlarge x limits = {value = .1},
    enlarge y limits={abs=0.8}
] 

\addplot[magenta!20!black,fill=magenta] coordinates {(5, 60.4) (7, 58.7) (9, 57.2) (11, 54.9)};
\addplot[cyan!20!black,fill=cyan] coordinates {(5, 59.9) (7, 57.7) (9, 56.7) (11, 54.9)};

%\addplot[magenta!20!black,fill=magenta] coordinates {(20, 8.9) (40, 9.1) (60, 8.9) (80, 8.5) (100, 8.3)};
%\addplot[cyan!20!black,fill=cyan] coordinates {(20, 1.7) (40, 1.9) (60, 1.2) (80, 1.7) (100, 1.7)};

\legend{MMC Transformer, Multiscale Baseline}

\end{axis}
\end{tikzpicture}
}%
\resizebox{0.5\textwidth}{!}{
\begin{tikzpicture}
\begin{axis} [
    width=\axisdefaultwidth,
    height=6cm,
    bar width = 5pt,
    ybar = .05cm,
    xmin = 0,
    xmax = 60,
    xtick={10,20,30,40,50},
    title = \textbf{High Frequencies},
    ylabel = mIoU,
    xlabel = $D_0$,
    area legend,
    enlarge x limits = {value = .1},
    enlarge y limits={abs=0.8}
] 

\addplot[magenta!20!black,fill=magenta] coordinates {(10, 19.3) (20, 17.7) (30, 16.5) (40, 14.8) (50, 13.9)};
\addplot[cyan!20!black,fill=cyan] coordinates {(10, 19.1) (20, 16.2) (30, 13.7) (40, 12.9) (50, 10.9)};

%\legend{MMC transformer, Baseline}

\end{axis}
\end{tikzpicture}
}
\caption{Evaluation of mIoU averaged over the four folds of our MMC transformer \vs the multiscale hypercorrelation squeeze network baseline~\cite{min2021hypercorrelation} with input images that have gone through band passed filtering. \textbf{Left:} Low pass filtering with different filter sizes and standard deviations. \textbf{Right:} High pass filtering with different cutoff frequencies using the complement of a butterworth low pass filter as $F=\frac{1}{1+\frac{D}{D_0}}$.}
\label{fig:freq_analysis}
\end{figure*}

\textbf{Frequency analysis.} We conduct additional ablations to study the robustness of our multiscale transformer decoding with multigrid exchange with respect to the different frequency components of our input signal. Inspired by previous work that analyzed transformers from a frequency domain perspective~\cite{bai2022improving}, we conduct a frequency analysis where we perform low pass filtering with a Gaussian kernel on the input images with different standard deviations and corresponding filter sizes.
%\textcolor{red}{TEXT IMPLIES THAT FIG. 2 WILL SHOW EXAMPLE FILTERED IMAGES, BUT INSTEAD IT SHOWS DATA PLOTS: FIX.} 
We then evaluate our final predictions similar to the original setup on the input filtered images and average per filter size over the four folds on YouTube-VIS. 
%The few-shot video segmentation task is already a challenging task on its own and when faced with such an additional difficulties the results degraded quite significantly for both the baseline and our approach; see Fig.~\ref{fig:freq_vis}. 
Fig.~\ref{fig:freq_analysis} (left) shows that for the three least low pass filtered cases, our approach better copes compared to the baseline; however, the advantage is lost at the most severe low pass filtering.  These results show that MMC transformer is more robust than the baseline to restricted frequency content, where here the restricted frequency content comes about through low pass filtering. 

In compliment, we conduct a similar study with the high frequency components restricted through application of the Fourier transform on our input signal, followed by high pass filtering with the complement of a butterworth low pass filter as $F=\frac{1}{1+\frac{D}{D_0}}$ with varying $D_0$ to control the cutoff frequency. Then we use the inverse Fourier transform and use the result as input to our models. 
%\textcolor{red}{IF YOU ARE GOING TO SHOW LOW PASS FILTERED IMAGES, AS SUGGESTED (BUT CURRENTLY NOT ACTUALLY SHOWN, THEN YOU NEED ALSO TO SHOW EXAMPLE HIGH-PASS FILTERED IMAGES AND IN DOING SO KEEP IN MIND ISSUES THAT APPEARED WHEN YOU PREPARED SUCH FOR NEURIPS SUBMISSION.} 
We report averaging over the four folds on YouTube-VIS. Figure~\ref{fig:freq_analysis} (right) shows that while both results are degraded by such filtering, our model degrades at a slower rate. These results further illustrate the robustness of our model to restricted frequency content. 
%\textcolor{red}{THE ABCISSA ON THE HIGH FREQUENCY PLOT SAYS "Filter Size", WHICH IS IRRELEVANT. INSTEAD NEED TO SHOW CUT-OFF FREQUENCY, AS THE CAPTION AND TEXT SUGGESTS. NOTE THAT THIS CHANGE LIKELY IS NOT JUST CHANGING THE LABEL; YOU MUST ALSO CORRESPONDINGLY CHANGE THE ACTUAL NUMBERS THAT ARE SHOWN ALONG THE AXIS. CURRENT DISPLAY IS INCORRECT!}

%These results confirm on our hypothesis that the multigrid formulation is more robust to band passed filtering especially with small filter sizes. %\textcolor{red}{Why do these results confirm the hypothesis?} 
%We leave it for future work to investigate the impact of low/high pass filtering on the few-shot task and how to design few-shot methods that are robust to such perturbations.

\textbf{Memory attention maps.} In the supplemental video, we visualise the attention maps from our multiscale memory decoding on the finest scale after the full bidirectional multigrid formulation is finalized for a video randomly selected from YouTube-VIS dataset. These visualizations allow us to gain better insights on how our multiscale memory learning enhances the output feature maps. Since we use $N=20$ we show the attention maps of the different memory entries along with the original image and the predicted segmentation (highlighted in red). It is seen that our multiscale memory decoding attends to different parts of the novel class and the background. It also shows the temporal consistency of the attended parts, for example memory entry 16 mostly attends to the novel class across all frames, memory entry three attends to parts of the background, while memory entry 11 attends to the hard negatives (\ie background pixels that are near pixels belonging to the novel class). These observations generally confirm the benefit from our multiscale memory decoding in enhancing the query feature maps and in separating the novel class from the background for better segmentation accuracy and using the temporal consistency nature within the input target videos. 

\begin{figure*}[t!]
\centering
\noindent
\begin{subfigure}{0.2\textwidth}
    \includegraphics[width=\textwidth]{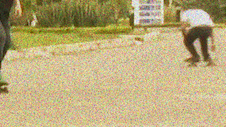}
\end{subfigure}%
\begin{subfigure}{0.2\textwidth}
    \includegraphics[width=\textwidth]{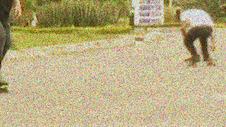}
\end{subfigure}%
\begin{subfigure}{0.2\textwidth}
    \includegraphics[width=\textwidth]{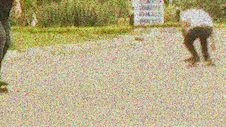}
\end{subfigure}%
\begin{subfigure}{0.2\textwidth}
    \includegraphics[width=\textwidth]{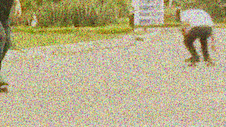}
\end{subfigure}%
\begin{subfigure}{0.2\textwidth}
    \includegraphics[width=\textwidth]{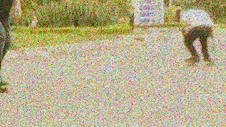}
\end{subfigure}

\begin{subfigure}{0.2\textwidth}
    \includegraphics[width=\textwidth]{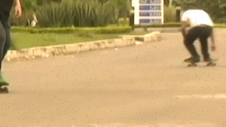}
    \caption{0.2}
\end{subfigure}%
\begin{subfigure}{0.2\textwidth}
    \includegraphics[width=\textwidth]{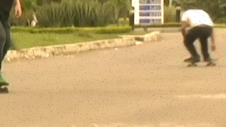}
     \caption{0.4}
\end{subfigure}%
\begin{subfigure}{0.2\textwidth}
    \includegraphics[width=\textwidth]{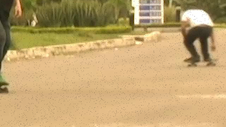}
     \caption{0.6}
\end{subfigure}%
\begin{subfigure}{0.2\textwidth}
    \includegraphics[width=\textwidth]{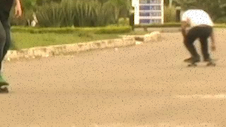}
    \caption{0.8}
\end{subfigure}%
\begin{subfigure}{0.2\textwidth}
    \includegraphics[width=\textwidth]{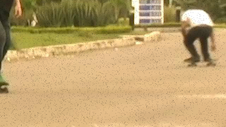}
     \caption{1.0}
\end{subfigure}
\caption{Visualization of speckle (top) and salt/pepper (bottom) noise with different configurations applied to an example image. In speckle noise we control the standard deviation of the Gaussian noise to range from 0.2 to 1.0. For the salt/pepper noise we control the percentage of pixels with values set to zero or one, with  percentage ranging from 0.2\% to 1\%.}
\label{fig:noise_imgs}
\end{figure*}

\textbf{Noise analysis full sweep.} We now present a fuller sweep across the range of noise corruptions for the cases of speckle and salt/pepper noise. We focus on these two and skip the additive Gaussian since both are closer to realistic scenarios that occur in medical image processing, \eg in ultrasound images where a snowstorm effect occurs and can degrade segmentation performance~\cite{noble2006ultrasound}. Additionally, the speckle noise already involves a Gaussian noise multiplied by the original signal. These results serve to augment the noise analysis presented in the main submission, which was limited to a single level of corruption for each type of noise. In this experiment we report the mean across folds on YouTube-VIS FS-VOS. 
%We pick the speckle since it encompasses a Gaussian noise and is closer to simulate the noise that can occur in ultrasound images~\cite{??}. 
In the speckle noise we control the standard deviation of the Gaussian multiplied by the original signal across the ranges 0.2 to 1.0. As a separate experiment, we manipulate the amount of salt/pepper noise added as the percentage of original pixel values set to zero or one, with the percentage ranging from 0.2 to 1\%. Figure~\ref{fig:noise_imgs} shows sample images perturbed by speckle noise (top) and salt/pepper noise (bottom). Note that the salt/pepper noise exhibits higher degree of randomness between runs since we not only randomly select the positions within the image to set to zero or one, we also randomly select the channel to set as well. For the sake of visualization only in Fig.~\ref{fig:noise_imgs}, we set all the channels to zero for both the salt and pepper noise since the salt when set to one is hard to see with light background.
%\textcolor{red}{LAST TWO SENTENCE CONFUSING. LAST SENTENCE: ARE YOU SAYING THAT THE PRESENTED EXAMPLE IMAGES DOES NOT SHOW THE ACTUAL CORRUPTION; THAT WOULD BE WIERD. PENULTIMATE SENTENCE: THE VISUALIZATION CONTRADICTS YOUR CLAIM, BECAUSE IT APPEARS THAT THE TOP ROW IS FAR MORE RANDOMLY CORRUPTED THAN THE BOTTOM ROW, WHICH THE OPPOSITE OF WHAT YOU TEXT SAYS. ALSO, THE FIGURE CAPTION IS INCORRECT, AS IT ONLY MENTIONS SALT AND PEPPER; CAPTION ALSO NEEDS TO EXPLAIN WHAT THE NUMBERS BETWEEN THE TWO ROWS OF FIGURES STAND FOR.}

\begin{figure*}[t!]
    \centering
\resizebox{0.45\textwidth}{!}{
\begin{tikzpicture}
\begin{axis} [
    width=\axisdefaultwidth,
    height=6cm,
    bar width = 5pt,
    ybar = .05cm,
    xmin=0,
    xmax=1.2,
    xtick={0.2,0.4,0.6,0.8,1.0},
    title = \textbf{Speckle Noise Analysis},
    ylabel = mIoU,
    xlabel = Noise,
    area legend,
    enlarge x limits = {value = .1},
    enlarge y limits={abs=0.8},
    legend columns=3,
    legend style={at={(0.5,-0.4)},anchor=south,row sep=0.01pt}
] 

\addplot[magenta!20!black,fill=magenta] coordinates {(0.2,48.8) (0.4,42.3) (0.6, 36.5) (0.8,32.5) (1.0,29.2)};
\addplot[orange!20!black,fill=orange] coordinates {(0.2,47.7) (0.4,41.7) (0.6, 35.3) (0.8,30.9) (1.0,28.3)};
\addplot[cyan!20!black,fill=cyan] coordinates {(0.2,47.3) (0.4,41.2) (0.6,35.7) (0.8,31.4) (1.0,27.7)};

\legend{MMemory-Multigrid, MMemory-Stacked, Baseline}

\end{axis}
\end{tikzpicture}
}%
\resizebox{0.45\textwidth}{!}{
\begin{tikzpicture}
\begin{axis} [
    width=\axisdefaultwidth,
    height=6cm,
    bar width = 5pt,
    ybar = .05cm,
    xmin=0,
    xmax=1.2,
    xtick={0.2,0.4,0.6,0.8,1.0},
    title = \textbf{Salt and Pepper Noise Analysis},
    ylabel = mIoU,
    xlabel = Noise,
    area legend,
    enlarge x limits = {value = .1},
    enlarge y limits={abs=0.8},
    legend columns=3,
    legend style={at={(0.5,-0.4)},anchor=south,row sep=0.01pt}
] 

\addplot[magenta!20!black,fill=magenta] coordinates {(0.2,61.9) (0.4,61.0) (0.6,60.3) (0.8,59.3) (1.0,59.1)};
\addplot[orange!20!black,fill=orange] coordinates {(0.2,61.0) (0.4,60.3) (0.6,60.2) (0.8,58.6) (1.0,58.5)};
\addplot[cyan!20!black,fill=cyan] coordinates {(0.2,60.9) (0.4,60.0) (0.6,60.2) (0.8,58.9) (1.0,58.7)};

\legend{MMemory-Multigrid, MMemory-Stacked, Baseline}

\end{axis}
\end{tikzpicture}
}
\caption{Evaluation of mIoU averaged over the four folds of our variants of multiscale memory learning (MMemory) \vs the conventional multiscale baseline~\cite{min2021hypercorrelation} with input images that have gone through different amounts of salt and pepper noise added to the original signal and different standard deviation for Gaussian noise multiplied then added to the original signal in speckle noise.}
\label{fig:noise_analysis}
\end{figure*}
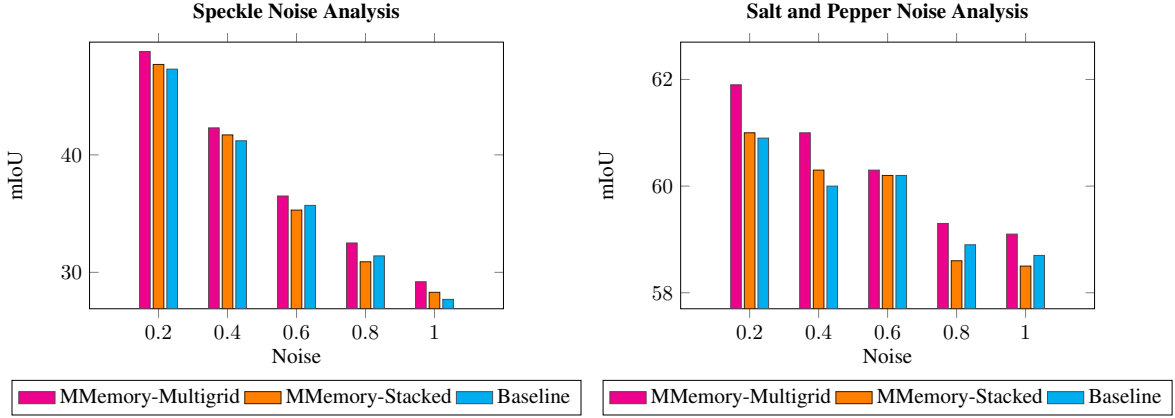

Figure~\ref{fig:noise_analysis} shows the result of the sweep for the speckle (left) and salt/pepper noise (right). 
%\textcolor{red}{THIS FIGURE IS APPEARING BEFORE THE PREVIOUS, BUT THEY NEED TO BE REVERSED TO MATCH THEIR ORDER OF INTRODUCTION IN THE TEXT.} 
Across the full range it is seen that as the noise severity increases the baseline and our variants decrease in performance according to mean intersection over union. Notably, however, our approach outperforms the multiscale baseline in the presence of both noise types. %the speckle noise case with mostly 1-2\% gain. 
Moreover, it is seen that our multigrid formulation is consistently more robust than the stacked information exchange across scales for both noise types.

\subsubsection{Additional qualitative results}
\begin{figure*}[t!]
\begin{subfigure}{0.5\textwidth}
\begin{subfigure}{0.33\textwidth}
    \includegraphics[width=\textwidth]{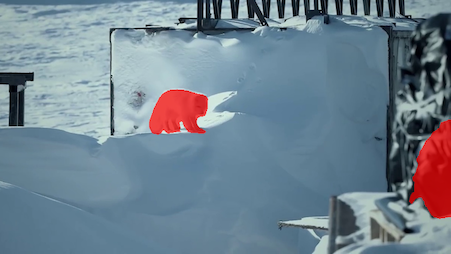}
\end{subfigure}%
\begin{subfigure}{0.33\textwidth}
    \includegraphics[width=\textwidth]{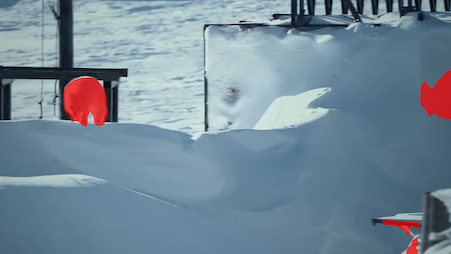}
\end{subfigure}%
\begin{subfigure}{0.33\textwidth}
    \includegraphics[width=\textwidth]{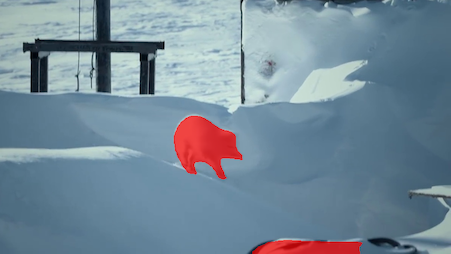}
\end{subfigure}
\end{subfigure}
\begin{subfigure}{0.5\textwidth}%
\begin{subfigure}{0.33\textwidth}
    \includegraphics[width=\textwidth]{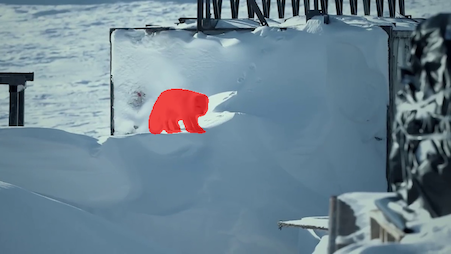}
\end{subfigure}%
\begin{subfigure}{0.33\textwidth}
    \includegraphics[width=\textwidth]{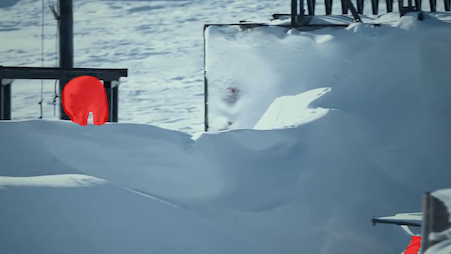}
\end{subfigure}%
\begin{subfigure}{0.33\textwidth}
    \includegraphics[width=\textwidth]{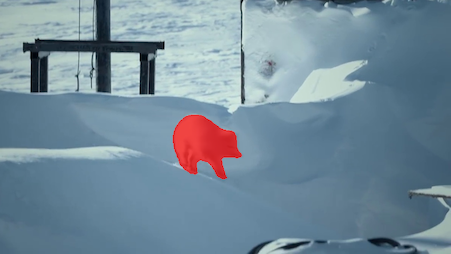}
\end{subfigure}
\end{subfigure}

\begin{subfigure}{0.5\textwidth}
\begin{subfigure}{0.33\textwidth}
    \includegraphics[width=\textwidth]{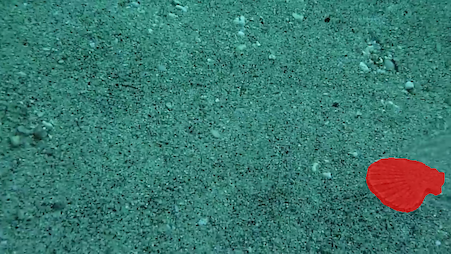}
\end{subfigure}%
\begin{subfigure}{0.33\textwidth}
    \includegraphics[width=\textwidth]{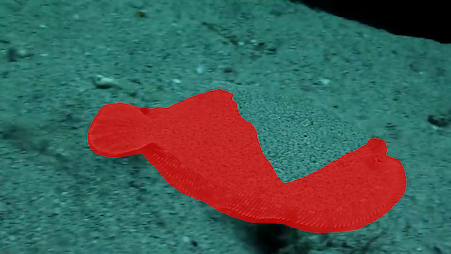}
\end{subfigure}%
\begin{subfigure}{0.33\textwidth}
    \includegraphics[width=\textwidth]{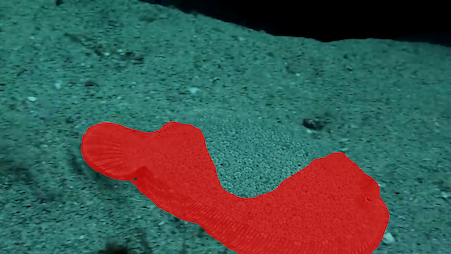}
\end{subfigure}
\end{subfigure}
\begin{subfigure}{0.5\textwidth}%
\begin{subfigure}{0.33\textwidth}
    \includegraphics[width=\textwidth]{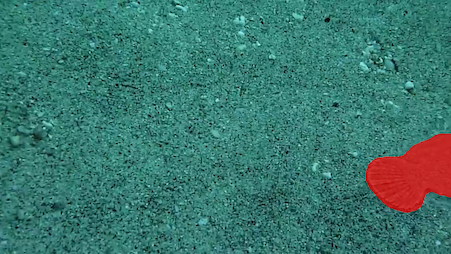}
\end{subfigure}%
\begin{subfigure}{0.33\textwidth}
    \includegraphics[width=\textwidth]{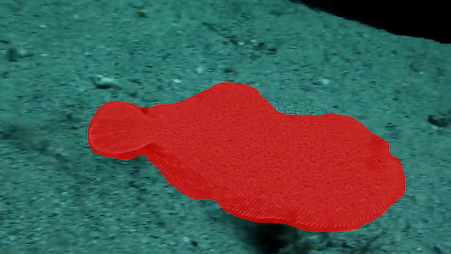}
\end{subfigure}%
\begin{subfigure}{0.33\textwidth}
    \includegraphics[width=\textwidth]{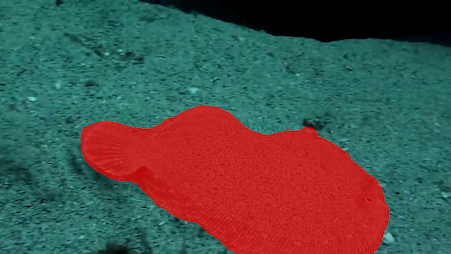}
\end{subfigure}
\end{subfigure}

\begin{subfigure}{0.5\textwidth}
\begin{subfigure}{0.33\textwidth}
    \includegraphics[width=\textwidth]{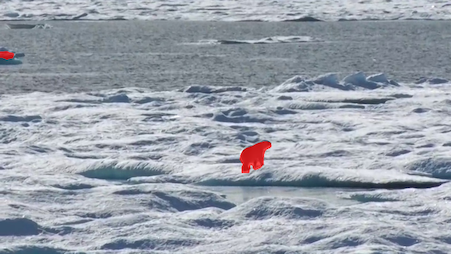}
\end{subfigure}%
\begin{subfigure}{0.33\textwidth}
    \includegraphics[width=\textwidth]{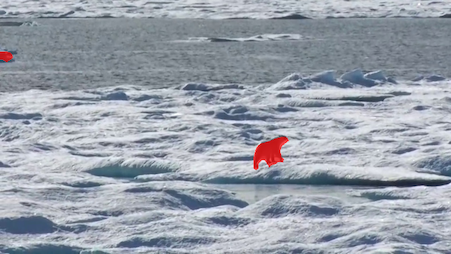}
\end{subfigure}%
\begin{subfigure}{0.33\textwidth}
    \includegraphics[width=\textwidth]{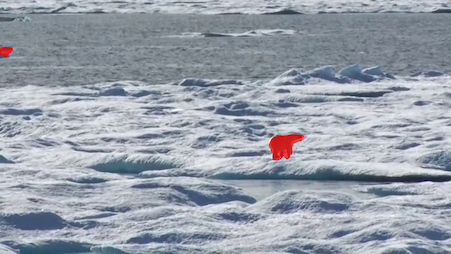}
\end{subfigure}
\caption{}
\end{subfigure}
\begin{subfigure}{0.5\textwidth}%
\begin{subfigure}{0.33\textwidth}
    \includegraphics[width=\textwidth]{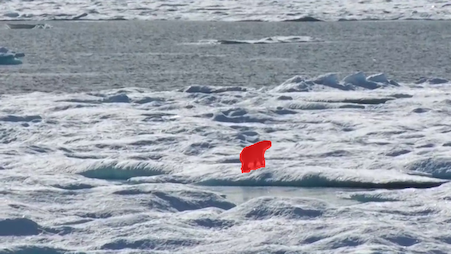}
\end{subfigure}%
\begin{subfigure}{0.33\textwidth}
    \includegraphics[width=\textwidth]{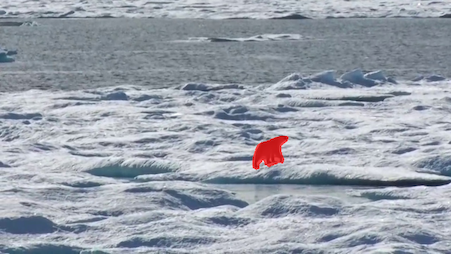}
\end{subfigure}%
\begin{subfigure}{0.33\textwidth}
    \includegraphics[width=\textwidth]{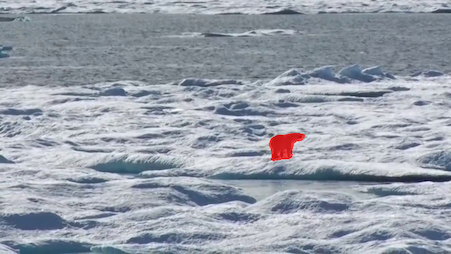}
\end{subfigure}
\caption{}
\end{subfigure}
\caption{Qualitative results on MoCA showing better segmentations for our multiscale multigrid memory decoding that exchanges information across scales on the detailed feature maps compared to our multiscale baseline that rather relies on a compact representation. (a) Multiscale Baseline. (b) Multiscale multigrid memory decoding (ours). Top row shows three frames from arctic fox video. Middle row shows three frames from flounder video. Bottom row shows three frames from polar bear video. Segmentation masks overlaid in red on each image.}
\label{fig:avos_qual}
\end{figure*}

In the main submission we presented qualitative results for  YouTube-VIS FS-VOS compared to the multiscale baseline. Here, we demonstrate this comparison to the multiscale baseline for fully supervised AVOS on three videos from MoCA. Figure~\ref{fig:avos_qual} shows that our approach is able to capture the camouflaged moving animals with varying sizes and challenges better than the conventional multiscale baseline that does not exchange information on the detailed feature maps. In the first row we show that for the arctic fox our baseline has multiple false positives that are considered as a foreground object, some of which are similar in appearance to the arctic fox. However, our method does not suffer from such false positives. In the middle row we show the camouflaged fish example where our method delineates the object boundaries better than the baseline. Finally, the last row shows the polar bear example, which similarly to the arctic fox, shows that our baseline exhibits false positives for confusing objects in the background, unlike our approach.

\subsection{Assets and licenses}\label{sec:assets}
We use the YouTube-VIS\footnote{\url{https://youtube-vos.org/dataset/}} dataset which is under a Creative Commons Attribution 4.0 License that allows non commercial research use. We use the DAVIS~\footnote{\url{https://davischallenge.org/davis2016/code.html}} and YouTube-VOS~\footnote{\url{https://youtube-vos.org/dataset/}} datasets during the training, where both are licensed under a Creative Commons Attribution 4.0 License that allows non-commercial research use. Additionally, we use YouTube Objects~\footnote{\url{https://data.vision.ee.ethz.ch/cvl/youtube-objects/}} and MoCA~\footnote{\url{https://www.robots.ox.ac.uk/~vgg/data/MoCA/}} for evaluation which are under the same license. We also used the A2D\footnote{\url{https://web.eecs.umich.edu/~jjcorso/r/a2d/}} dataset where the license states that the dataset may not be republished in any form without the written consent of the authors. %We rely on certain parts from the code of Vis-TR to create our baseline 

\subsection{Societal impact}\label{sec:societalimpact}
Few-shot video object segmentation, where the query set to be segmented is a video, is a crucial task that can help reduce the annotation cost required to label large-scale video datasets. It can serve a variety of applications in autonomous systems~\cite{cen2021deep} and medical image processing~\cite{ross2020robust} which require the model to learn from few labelled examples for novel classes that are beyond the closed set of training classes with abundant labels. It can also help bridge the gap between developing and developed countries, where the former lacks the resources necessary to annotate large-scale labelled datasets that are required in a variety of tasks that serves the community such as, the use of satellite imagery in agricultural monitoring and crop management~\cite{segarra2020remote}. We believe our work in general provides positive impact in empowering developing countries to establish labelled datasets that satisfy the needs of their own communities rather than following public benchmarks.

However, as with many artificial intelligence algorithms, video object segmentation can have negative societal impacts, \eg through application to automatic target detection in military and surveillance systems. There are emerging movements to limit such applications, \eg  pledges on the part of researchers to band use  of artificial intelligence in weaponry systems. We have participated in signing that pledge and are supporters of its enforcement through international laws. Nonetheless, we strongly believe these misuses are available in both few-shot and non few-shot methods and are not tied to the specific few-shot case. On the contrary, we argue that empowering developing countries towards decolonizing artificial intelligence can help go beyond centered power that currently lies within developed countries and big tech companies and is guided solely by their interests.

{\small
\bibliographystyle{ieee_fullname}
\bibliography{egbib}
}

\end{document}